\documentclass[10pt,journal,compsoc]{IEEEtran}
\ifCLASSOPTIONcompsoc
  \usepackage[nocompress]{cite}
\else
  \usepackage{cite}
\fi
\ifCLASSINFOpdf
\else
\fi
\usepackage{stfloats}
\usepackage{float}
\usepackage{comment}
\usepackage{graphicx}
\usepackage{makecell}
\usepackage{url}
\usepackage{ragged2e}
\usepackage{amsmath,amssymb}
\usepackage{utfsym}
\usepackage[pagebackref,breaklinks,colorlinks]{hyperref}
\usepackage[ruled,linesnumbered]{algorithm2e}
\usepackage{lipsum} 
\usepackage{multicol} 
\usepackage{arydshln}

\newcommand{\customtitle}[2]{%
  \twocolumn[
    \begin{@twocolumnfalse}
      \begin{center}%
        \vspace{1em}
        \Huge \bfseries #1 \par 
        \vspace{1em} 
      \end{center}%
      #2 
    \end{@twocolumnfalse}
}
\begin{document}
%
\title{InfoBFR: Real-World Blind Face Restoration via Information Bottleneck} 

%
%
%

\author{Nan Gao$^{\dagger}$,
        Jia Li$^{\dagger}$,
        Huaibo Huang,
        Ke Shang, 
        Ran He$^{*}$, IEEE Fellow
\IEEEcompsocitemizethanks{\IEEEcompsocthanksitem Nan Gao and Jia Li contributed equally to this work. Nan Gao is with the Institute of Automation Chinese Academy of Sciences, Beijing,
China.\protect\\
E-mail: nan.gao@ia.ac.cn
\IEEEcompsocthanksitem Jia Li and Ke Shang are with PCIE, Lenovo Research.
\IEEEcompsocthanksitem Huaibo Huang and Ran He (Corresponding author) are with the New Laboratory of Pattern Recognition, CASIA.}}

\IEEEtitleabstractindextext{%

\begin{abstract}
\justifying
Blind face restoration (BFR) is a highly challenging problem due to the uncertainty of data degradation patterns. Current BFR methods have realized certain restored productions but with inherent neural degradations that limit real-world generalization in complicated scenarios. In this paper, we propose a plug-and-play framework \emph{InfoBFR} to tackle neural degradations, e.g., prior bias, topological distortion, textural distortion, and artifact residues, which achieves high-generalization face restoration in diverse wild and heterogeneous scenes. Specifically, based on the results from pre-trained BFR models, InfoBFR considers information compression using manifold information bottleneck (MIB) and information compensation with efficient diffusion LoRA to conduct information optimization. InfoBFR effectively synthesizes high-fidelity faces without attribute and identity distortions. Comprehensive experimental results demonstrate the superiority of InfoBFR over state-of-the-art GAN-based and diffusion-based BFR methods, with around 70ms consumption, 16M trainable parameters, and nearly \textbf{85\%} BFR-boosting. It is promising that InfoBFR will be the first plug-and-play restorer universally employed by diverse BFR models to conquer neural degradations.
\end{abstract}

}

\maketitle

\IEEEdisplaynontitleabstractindextext

%
\IEEEpeerreviewmaketitle

\IEEEraisesectionheading{\section{Introduction}\label{sec:introduction}}

%
%
%
%

\begin{figure}[htbp]
\begin{center}
   \includegraphics[width=1\linewidth]{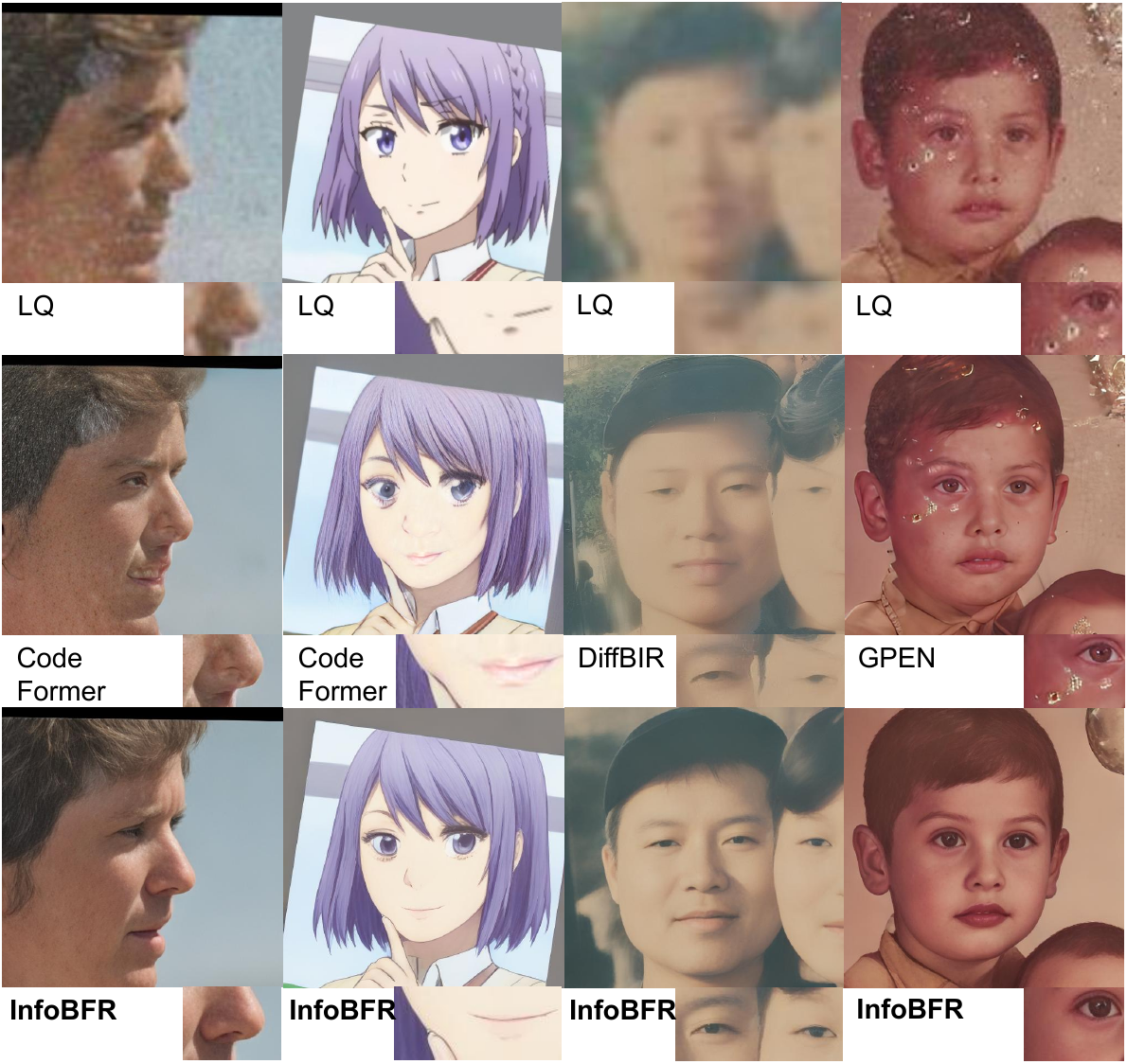}
\end{center}
   \caption{InfoBFR helps pre-trained BFR models (CodeFormer \cite{codeformer}, DiffBIR \cite{diffbir}, GPEN \cite{GPEN}) fight against neural degradations, e.g., topology distortion (col 1), prior bias (col 2), texture distortion (col 3), and artifact residues (col 4). Note that pre-trained BFR models and our proposed InfoBFR are both trained on the FFHQ \cite{style2019} dataset. More visual examples are shown in Fig. \ref{fig:stage2}.}
\label{fig:start}
\end{figure}

\IEEEPARstart{R}{eal-world} blind face restoration is a challenging task, particularly when dealing with severe data degradations and out-of-domain scenes (Fig. \ref{fig:start}) for current BFR neural models. As a training dataset protocol, FFHQ \cite{style2019} mainly contains photorealistic human faces, based on which several recent state-of-the-art models synthesize high-quality BFR results. Specifically, GAN-based methods \cite{vqfr, codeformer, GPEN, gfp2021, femasr, dfdnet2020} and diffusion-based methods \cite{difface, diffbir} have achieved acknowledged BFR performance, which leverages outstanding feature representation ability of pre-trained models, such as VQ-GAN \cite{vqgan}, StyleGAN \cite{style2019}, and Stable Diffusion \cite{latentdiff}.


  


To some extent, in some simple real-world scenarios such as LFW \cite{lfw} dataset with mild data degradations, these BFR models have successfully removed the data degradations such as Gaussian blurring, motion blurring, image noise, JPEG compaction, and imaging artifacts.


\begin{figure*}[htbp]
\begin{center}
   \includegraphics[width=1\linewidth]{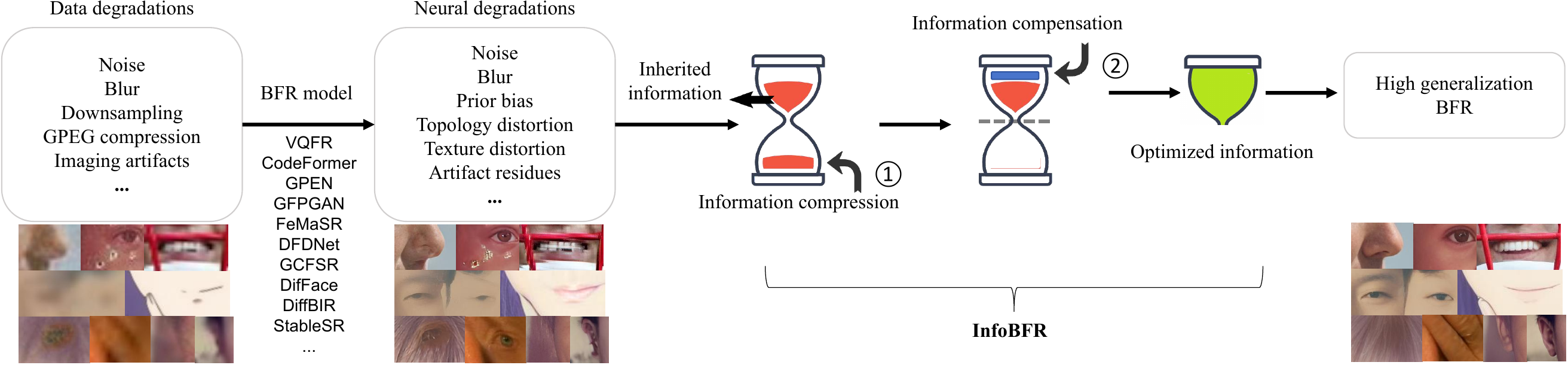}
\end{center}
   \caption{Motivation of InfoBFR. In low-quality data distribution, diverse data degradations are caused by physical reasons such as optical imaging of images, external contamination (col 4, Fig \ref{fig:start}), or digital processing reasons such as image transmission, image enc-dec. Low-generalization BFR models suffer from severe data degradations and produce high-quality images to a certain extent. However, neural BFR models give birth to new problems, i.e., neural hallucinations during model inference. We define these urgent and neglected degradations as \emph{\textbf{neural degradations}}. We first study the universal neural degradation restoration method \emph{\textbf{InfoBFR}}. Specifically, we conduct information compression and compensation where novel details and inherited information from the pre-trained BFR model are neurally integrated. }
\label{fig:moti}
\end{figure*}

However, in some complicated real-world scenarios such as WebPhoto \cite{gfp2021}, these methods are not powerful enough even bring additional image degradations that we call \textbf{neural degradations} which means accessory degradations of neural networks, as shown in Fig. \ref{fig:moti}.  Moreover, as shown in Fig. \ref{fig:start}, we give some visual examples of prior bias (col 2), topology distortion (col 1), texture distortion (col 3), and artifact residues (col 4). It is essential to "\emph{absorb the essence and discard the useless}", i.e., only undistorted features are supposed to be inherited from the pre-trained BFR models.

So the question is how to polish these pre-trained BFR models to conquer the neural degradations. To this end, we propose a novel information optimization strategy called \emph{InfoBFR}. The key idea is as shown in Fig. \ref{fig:moti}, there are two neural steps concerning information compression and information compensation. Concretely, first, neural degradations are supposed to be removed. Second, more high-quality facial details are required to be synthesized for better topological and textural distribution based on the first step. 

Inspired by information compression learning \cite{ib, iba} designed to filter out noised information and extract valuable information for specific tasks, we explore leveraging information bottleneck (IB) \cite{iba} for the first step, i.e.,  optimizing a trade-off between information purification and preservation based on relevant information theory. An effective IB is designed to maximize the mutual information (MI) between the optimized and real manifold while minimizing MI between the pre-trained BFR and optimized manifold. Model details of our approach are in Section \ref{section:app}.


Diffusion model has been potentially used on high-quality image syntheses \cite{latentdiff, controlnet, diffbir}. The latent diffusion model \cite{latentdiff} manipulates perceptual image compression to the manifold space followed by the latent diffusion modeling. Manifold representation learning is equivalent to image-level representation learning with the pre-trained VAE. Therefore, in our proposed \emph{InfoBFR}, we design the information bottleneck and diffusion compensation on the manifold level. Furthermore, we apply the LoRA \cite{lora} strategy to facilitate the denoising process in only one step. 

We explore discovering a universal approach to equip diverse BFR models with a strong ability to defend against neural degradations. As a result, InfoBFR can effectively synthesize high-fidelity faces for diverse scenes. Qualitative improvement results are shown in Fig. \ref{fig:stage2}, \ref{fig:extreme}, and the supplementary materials.

\begin{figure}[tbp]
\begin{center}
   \includegraphics[width=1\linewidth]{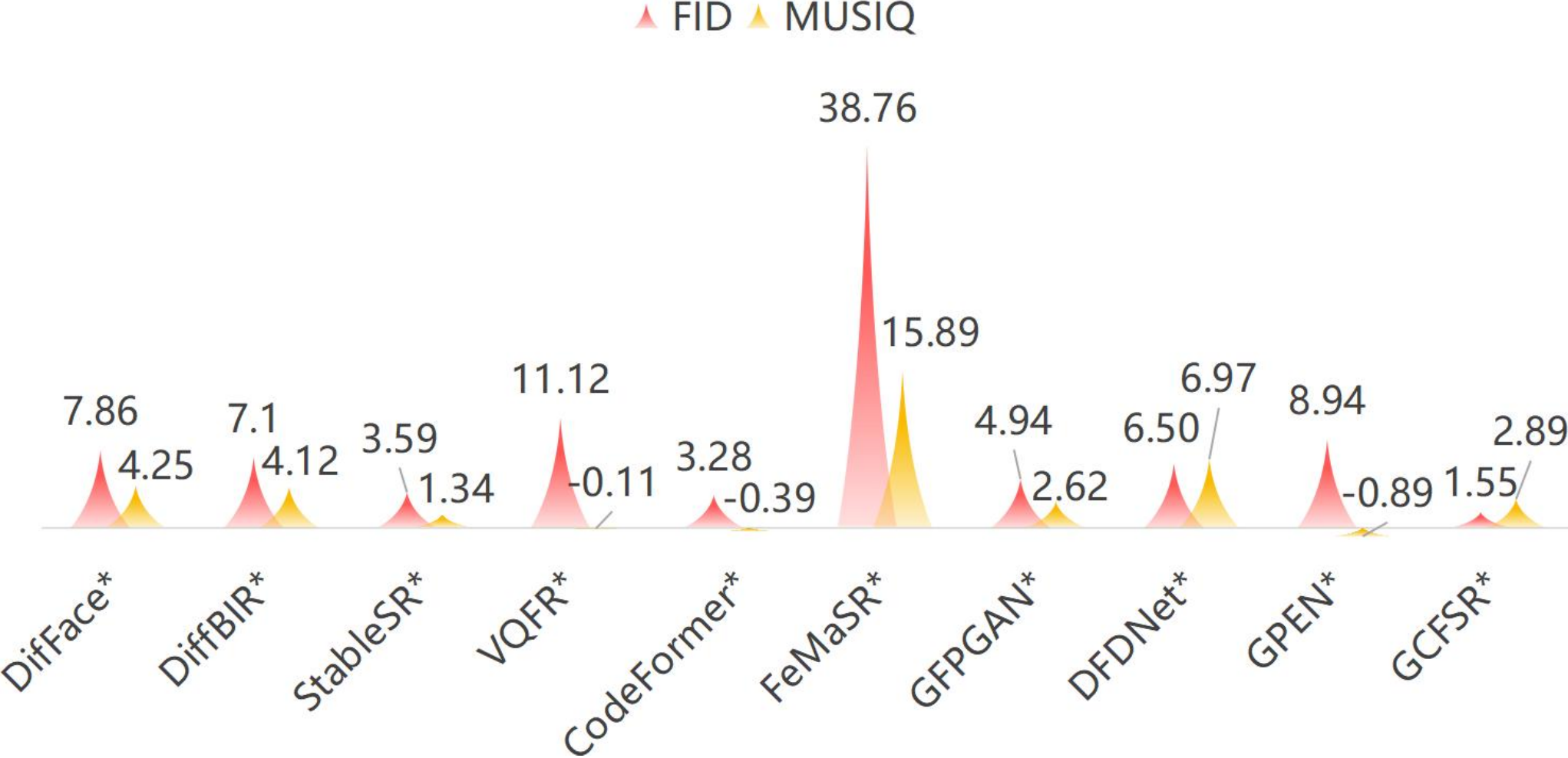}
\end{center}
   \caption{Comprehensive FID-boosting and MUSIQ-boosting visualization of BFR* models equipped with plug-and-play InfoBFR, compared with the original BFR in real-world datasets with heavy degradations (Wider-Test \cite{codeformer}, FOS \cite{fos}, CelebChild \cite{gfp2021} and WebPhoto \cite{gfp2021}). InfoBFR facilitates BFR quality by effectively harnessing the information bottleneck. More quantitative evaluations are shown in Tab. \ref{tab:improve} and Tab. \ref{tab:real}.}
\label{fig:compare}
\end{figure}
Our paper presents several significant contributions mainly including three folds:
\begin{itemize}
\item We propose a high-generalization BFR framework called InfoBFR to conquer the challenging neural degradations, while almost all BFR models focus on data degradations. Considering information compression and compensation, our approach achieves exceptional BFR results for real-world LQ images with severely degraded scenes. 

\item We present an efficient and effective Manifold Information Bottleneck (MIB) module that provides a trade-off between diffusion manifold preservation and compression. MIB functionally disentangles inherited manifold and neural degradations, which improves the restoration controllability. Moreover, one-step diffusion compensation further builds up restoration fidelity and quality.

\item We first study the challenging neural degradation restoration problem for the real-world BFR task. Compared with state-of-the-art model-based and dictionary-based approaches,  comprehensive experimental analyses demonstrate that InfoBFR is the first effective plug-and-play restorer for pre-trained BFR models with nearly \textbf{85\%} BFR-boosting quantitative items (Tab \ref{tab:real}, Tab \ref{tab:improve}).
\end{itemize}

\section{Related Work}
\label{gen_inst}
There are mainly three kinds of methods that leverage the pre-trained model, facial prior of the reference face, or high-quality feature bank to conduct BFR. We introduce these works and information compression works in this section.
\subsection{Model-based Image Restoration} 
Several remarkable methods that leverage pre-trained StyleGAN \cite{stylegan2_2020} as the face prior have been proposed recently, including GLEAN \cite{bank2021}, GFP-GAN \cite{gfp2021}, PULSE \cite{pulse2020}, GPEN \cite{GPEN}, et al. Another kind of promising approach is based on a stable diffusion model, such as DifFace \cite{difface}, and DiffBIR \cite{diffbir} that contains two stages respectively for degradation removal based on SwinIR and denoising refinement based on diffusion mechanism. Note that our InfoBFR also gives the diffusion model an important role.
\begin{figure*}[htbp]
\begin{center}
   \includegraphics[width=1\linewidth]{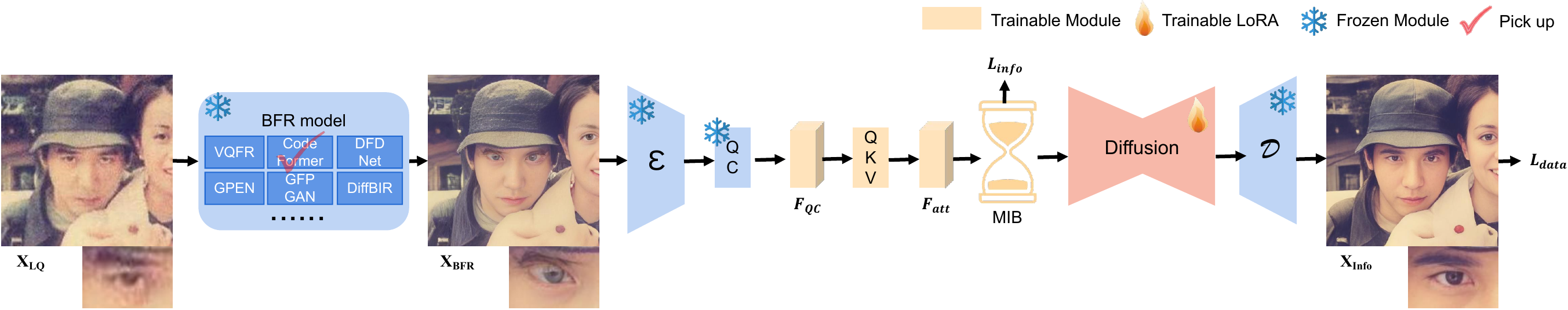}
\end{center}
   \caption{InfoBFR framework adopts a diffusion finetuning strategy to tackle the challenging real-world BFR task, novelly encompassing transformer, MIB, and diffusion LoRA. Fig. \ref{fig:ablation} and Table \ref{tab:ablation} demonstrate the indispensability of transformer and MIB. We implement manifold information compression from neurally distorted manifold $\mathcal{E}_{X_{BFR}}$ followed by a QKV attention. Based on compressed feature $Z$ in Algorithm \ref{alg:alg1}, diffusion LoRA is then used to control feature transformations for real-world BFR adaptively. QC means $quant\_conv$ layer of the encoder of pre-trained VAE. Overall, the InfoBFR framework effectively achieves high-quality BFR results with robustness in diverse domains. Powerful supporting experiments are shown in Table \ref{tab:real} and figures of the paper and supplementary material.}
\label{fig:pipeline}
\end{figure*}
\subsection{Reference-based Image Restoration} 
Reference-based methods with guided images have been proposed as well, such as WarpNet \cite{warpnet2018}, ASFFNet \cite{exemplar2020}, CIMR-SR\cite{pool2020}, CPGAN \cite{copy2020}, Masa-sr \cite{masa2021}, PSFRGAN \cite{psfr}. These methods explore and incorporate the priors of reference images,  e.g., facial landmarks, identity, semantic parsing, or texture styles, for adaptive feature transformation. Our InfoBFR conducts information bottleneck based on the facial prior of the pre-trained BFR models.

\subsection{Bank-based Image Restoration}
These methods first establish high-quality feature dictionaries based on VQGAN \cite{vqgan} or VQVAE \cite{vqvae}, followed by feature matching for image restoration, for example, DFDNet \cite{dfdnet2020}, Codebook Lookup Transformer (CoLT) \cite{codeformer},  VQFR \cite{vqfr}. In contrast, our InfoBFR operates restoration without any prior bank.
\subsection{Information Compression}
Tishby \cite{ib} first proposes an information bottleneck (IB)  that takes a trade-off between information compression and robust representation ability for specific tasks. VIB \cite{vib} leverages the reparameterization trick \cite{trick} with variational approximation to train the IB neural layer efficiently. IBA \cite{iba} restricts the attribution information using adaptive IB for effective disentanglement of classification-relative and irrelative information. InfoSwap \cite{infoswap} views face swapping in the plane of information compression to generate identity-discriminative faces. Furthermore, Yang \cite{infopaint} proposes to use a highly compressed representation that maintains the semantic feature while ignoring the noisy details for better image inpainting. We will introduce our manifold information bottleneck in Section \ref{section:mib}.

\section{Approach}
\label{section:app}
\subsection{Problem modeling}
Revisit that the denoising diffusion probabilistic models (DDPMs) \cite{denoising, diffusion, ddpm} take care of the diffusion modeling with noise regularization, while ignoring the denoising process regularization aligned to the image-level ground truth. Unfortunately, this manner is weak to provide strong perceptual imaging constraints for controllable high-fidelity image syntheses, especially in some visually sensitive areas, e.g., teeth. Another problem is the multiple random denoising steps that not only consume too much time (Tab. \ref{tab:time}) but also introduce more uncertainty. Existing BFR methods, e.g., DifFace \cite{difface}, DiffBIR \cite{diffbir}, and StableSR \cite{stablesr}, heavily suffer from these ineffective training processes.

The good news is that the pre-trained BFR model has pushed the BFR process to the altitude of the approaching finish line which can be taken as the beginning point for further denoising optimization of DDPMs. As shown in Fig \ref{fig:moti}, InfoBFR is designed to fight against neural degradations. In the information compensation step, we favorably conduct only one-step fine-tuned denoising with LoRA adoption to eliminate the image-level regulation shortage and computational cost. We define our neural degradation restoration learning as:
\begin{equation}
\label{equ:g}
\begin{aligned}
\theta^{*} &=argmin_{\theta}\mathbb{E}_{X_{BFR},X_{GT}}[\mathcal{L}_{data}(G_{\theta}(X_{BFR}),X_{GT})\\
&+\mathcal{L}_{info}(MIB_{\theta}(X_{BFR}))],
\end{aligned}
\end{equation}
where $G_{\theta}$ is the generator of InfoBFR, $X_{BFR}$ is the BFR results from GAN-based or Diffusion-based models, $X_{GT}$ is the ground-truth. $\mathcal{L}_{data}$ aims to provide data-level regularization for the optimized information in Fig. \ref{fig:moti}, and $\mathcal{L}_{info}$ seeks to control feature neural selection namely information compression. Both are beneficial to stay away from neural degradations, that is, defective genes inherited from the mother BFR model.

More importantly, we leverage information bottleneck to address the neural degradations that are dangerous obstacles to reaching the BFR peak's end. Let's denote the original input data, the corresponding label, and compressed information by $X$, $Y$, and $Z$. In information bottleneck concept \cite{ib, ib_concept}, the information compression principle is a trade-off between information preservation and the precise representation capability aligned with the supervising signal, that is, maximizing sharable information of $Z$ and $Y$ while minimizing sharable information of $Z$ and $X$:

\begin{equation}
\label{equ:mib}
\mathop{\max}_{Z} \mathbb{I}(Y;Z) -\beta\mathbb{I}(X;Z),
\end{equation}
where $\mathbb{I}$ means the mutual information and $\beta$ is trade-off weight.  Let $R$ indicate the intermediate representations of $X$. The equivalent definition of $\mathbb{I}(X;Z)$ is:

\begin{equation}
\label{equ:3}
\mathbb{I}(X;Z) \triangleq{\mathbb{I}(R;Z)} \triangleq{\mathcal{D}_{KL}[p(Z|R)\Vert q(Z)]},
\end{equation}
which is the information loss function where $q(Z)$ with Gaussian distribution is a variational approximation of $p(Z)$ \cite{iba}. $\mathcal{D}_{KL}$ means the KL divergence \cite{kl} used to measure the distance between two distributions. KL divergence helps VAE and GAN with mathematical modeling, especially promising for Gaussian distribution.

In our problem modeling, information loss is formulated as:
\begin{equation}
\mathcal{L}_{info} =\mathcal{D}_{KL}[p(\small MIB_{\theta}(X_{BFR})|\mathcal{F}_{R}(X_{BFR}))\|q(MIB_{\theta}(X_{BFR}))],
\end{equation} 
where $\mathcal{F}_{R}$ is the manifold process function including Encoder $\mathcal{E}$ and quant\_conv layer of VAE \cite{latentdiff} followed by a Transformer layer.

Meanwhile, according to Equ. \ref{equ:g}, $\mathbb{I}(Y;Z)$ can be accessed as follows:

\begin{equation}
\mathbb{I}(Y;Z) \triangleq{\mathbb{I}(X_{GT};G_{\theta}(X_{BFR}))}.
\end{equation}

\subsection{InfoBFR}
\label{section:InfoBFR}
In this section, we provide a detailed introduction to our proposed InfoBFR method, including the overall InfoBFR framework in Sec. \ref{section:frame}, manifold information bottleneck (MIB) module in Sec. \ref{section:mib}, along with the training loss in Sec. \ref{section:loss}. 
\subsubsection{Framework overview}
\label{section:frame}
Latent diffusion model \cite{latentdiff} conducts diffusion process on the compressed latent, i.e., manifold of the image distribution. This efficient manner is also leveraged in ControlNet \cite{controlnet}, StableSR \cite{stablesr} and DiffBIR \cite{diffbir}. The distribution constraint of the latent diffusion model is:
\begin{equation}
\mathcal{L}_{ldm}=\mathbb{E}_{z,c,t,\epsilon}[\Vert\epsilon-\epsilon_{\theta}(z_{t}=\sqrt{\overline{\alpha}_{t}}z+\sqrt{1-\overline{\alpha}_{t}}\epsilon, c, t)\Vert_{2}^{2}],
\end{equation},
where $z$ denotes the manifold obtained via encoder of VAE, i.e., $z=\mathcal{E}(I_{HQ})$.
$\epsilon \sim \mathcal{N}(0,\mathbb{I})$ with variance $\beta_{t}=1-\alpha_{t}\in(0,1)$ that is used to generate noisy manifold.

Once the optimization of the latent diffusion model is finished, the denoised manifold is calculated as follows:
\begin{equation}
\Tilde{z}_{0}=\frac{z_{t}}{\sqrt{\overline{\alpha}_{t}}}-\frac{\sqrt{1-\overline{\alpha}_{t}}\epsilon_{\theta}(z_{t},c,t)}{\sqrt{\overline{\alpha}_{t}}}.
\label{eq3}
\end{equation}

Given a target face $X_{HQ}$, we obtain the degraded image $X_{LQ}$ using sequential degradation methods such as blurring, downsampling, noise injection, and JPEG compression. As shown in Figure \ref{fig:pipeline}, we first transfer $X_{LQ}$ to $X_{BFR}$, then implement manifold compression to make distorted manifold $\mathcal{E}_{X_{BFR}}$ stay away from the neural degradations based on information bottleneck. Subsequently, we equip the pre-trained Unet of SD model with multi-scale finetuned features $z_{S}=\{z_{S}^{1}, z_{S}^{2}, ..., z_{S}^{n}\}$ based on LoRA \cite{lora} and impose the pre-trained diffusion model to compensate useful information towards to the target facial distribution adaptively. 

By denoting the trainable fine-tuned parameters as $\Delta W^{i}_{LoRA}$, and the frozen pretrained weights as $W^{i}_{\phi}$, we have the re-target updated parameters as follows:
\begin{equation}
W^{i}_{\theta}=W^{i}_{\phi} + \Delta W^{i}_{LoRA}.
\label{eq3}
\end{equation}

As an end-to-end model in Fig \ref{fig:pipeline}, InfoBFR synthesizes output as:
\begin{equation}
\begin{aligned}
{X}_{Info} &=G_{\theta}({X_{BFR}})\\
           &\triangleq{\mathcal{D}[LoRA(MIB(\mathcal{F}_{Att}(QC(\mathcal{E}(\mathcal{F}_{BFR}(X_{LQ}))))))]},
\end{aligned}
\end{equation} 
where $X_{BFR}=\mathcal{F}_{BFR}(X_{LQ})$, $\mathcal{F}_{R} = \mathcal{F}_{Att}(QC(\mathcal{E}))$, $\mathcal{D}$ is the decoder of VAE \cite{latentdiff}. Note that we only finetune the UNet denoising module while maintaining VAE parameters ($\mathcal{E}$, QC, $\mathcal{D}$) frozen, assuming $X_{BFR}$ and ${X}_{Info}$ share the same data distribution as real faces. Algorithm pseudo code is described in Algorithm \ref{alg:alg1}.

\begin{figure*}[htbp]
\begin{center}
   \includegraphics[width=0.9\linewidth]{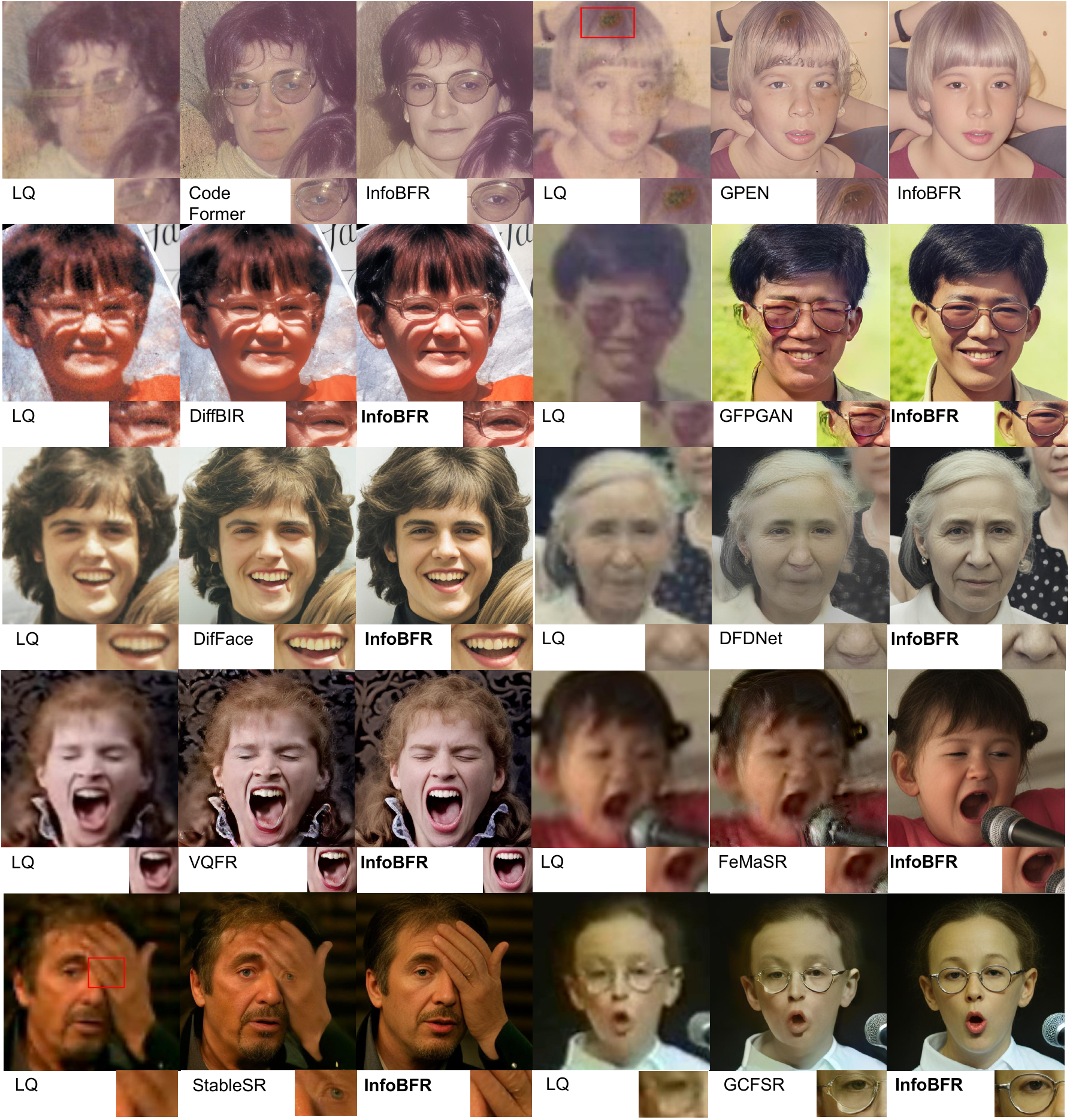}
\end{center}
   \caption{InfoBFR plays an important role in severely degraded scenarios. BFR models equipped with InfoBFR demonstrate more natural and cleaner than original BFR models beleaguered by neural degradations such as facial noises, ambiguous textures, external contamination ( pollution artifacts in Row 1), and prior bias (Row 5). Zoom in for better visual observation. More results are shown in Fig. \ref{fig:extreme} and the supplementary material.}
\label{fig:stage2} 
\end{figure*}

\subsubsection{Manifold Information Bottleneck}
\label{section:mib}
MIB is designed as a powerful function that maximally compresses the neural degradations caused by the pre-trained BFR model. MIB can be formulated as follows:
\begin{equation}
\label{equ:mib}
\mathop{\min}_{Z} \beta\mathbb{I}(R;Z)
\end{equation},
where $\mathbb{I}$ means the mutual information function, $Z$ is the optimal representation based on $R$ in line 4 of Algorithm \ref{alg:alg1}.

Note that we introduce a Transformer block $\mathcal{F}_{Att}$ to impose $R$ has a spacial-perception view for MIB with neural attention. 

\begin{equation}
\label{equ:att}
\mathcal{F}_{Att} = Softmax(\frac{Q\cdot K}{\sqrt{d}})V, 
\end{equation}
where ${Q,K,V}$, i.e., ${query, \ key, \ value}$ sets are from $QC(\mathcal{E}(X_{BFR}))$, $d$ is used to stabilize the gradient. 

\begin{algorithm}[tbp]
\caption{Algorithm concerning InfoBFR}\label{alg:alg1}
\SetKwInOut{Input}{Input}\SetKwInOut{Output}{Output}
\SetKwInOut{Require}{Require}
\Input{$X_{LQ}$, $X_{HQ}$, text prompt $c$ (set to " ")}
\Require{Pretrained BFR model $\mathcal{F}_{BFR}$, Latent diffusion model $\epsilon_{\phi}$ (Stable diffusion v2.1), VAE's encoder $\mathcal{E}$, quant\_conv $QC$ and decoder $\mathcal{D}$, transformer $\mathcal{F}_{Att}$}
\Require{Pre-calculated mean $\mu_{QC}$ and std. $\sigma_{QC}$ values based on a small dataset of $X_{HQ}$}
\Output{Restored face $X_{Info}$}

$\mathcal{L}_{info}\leftarrow 0$, $\mathcal{L}_{data}\leftarrow 0$\;%
$X_{BFR} = \mathcal{F}_{BFR}(X_{LQ})$\;
$R\leftarrow QC(\mathcal{E}(X_{BFR}))$\; 
$R\leftarrow \mathcal{F}_{Att}(R)$;\hfill{$\triangleright \ $Transformer}\ 

$\mathbb{F}_{R}\leftarrow \frac{R-\mu_{QC}}{max(\sigma_{QC},o)}$\;
$\lambda \leftarrow Sigmoid(Conv(\mathbb{F}_{R}))$\;
$Z\leftarrow \lambda*R$;\hfill{$\triangleright \ $MIB}\ 

$\mathcal{L}_{info}\leftarrow \mathbb{I}[Z,R]$\;
$Z\leftarrow $ DiagonalGaussianDistribution($Z$)\;

Initialize $ \epsilon_{\theta} \leftarrow \epsilon_{\phi}$ with trainable LoRA\;
Sample $\Tilde{z}_{0}\leftarrow \frac{Z}{\sqrt{\overline{\alpha}}}-\frac{\sqrt{1-\overline{\alpha}}\epsilon_{\theta}(Z,c)}{\sqrt{\overline{\alpha}}}$;\hfill{$\triangleright \ $Diffusion LoRA}\

$X_{Info}\leftarrow \mathcal{D}(\Tilde{z}_{0})$.\hfill{$\triangleright \ $Decoder}\

$\mathcal{L}_{data}\leftarrow \mathbb{I}[X_{Info},X_{HQ}]$\;
\end{algorithm}

To extract the useful manifold and stay away from the noised manifold with neural degradations, we design an information filter $\lambda$. Given $R \sim \mathcal{N}(\mu_{\mathcal{M}},\sigma_{\mathcal{M}}^{2})$, 
 where $\mu_{\mathcal{M}}$ and $\sigma_{\mathcal{M}}$ represent the means and standard deviations of $R$. We assume $R$ shares the same distribution with the real-data distribution, so $R \sim \mathcal{N}(\mu_{\mathcal{QC}},\sigma_{\mathcal{QC}}^{2})$. 

Then, the optimized manifold can be formulated as follows \cite{iba}:
\begin{equation}
Z = \lambda R+(1-\lambda)\epsilon
\label{equ:compress}
\end{equation},
where $Z$, $R$ and  $\epsilon$ are from $\mathcal{N}(\mu_{\mathcal{M}},\sigma_{\mathcal{M}}^{2})$ since we set $Z$ and $R$ have a consistent distribution. The goal of InfoBRF is to remove neural degradations of $X_{BFR}$ while adhering to the global facial attributes of $X_{BFR}$. If $\lambda$ is 0, the whole manifold will be replaced by random Gaussian noise $\epsilon$, which results in uncontrollable face restoration. If $\lambda$ is 1, neural degradations will not be eliminated. Therefore, information filter $\lambda$ plays a crucial role in MIB. Qualitative analyses are illustrated in Fig \ref{fig:ablation}.

After manifold information compression, we obtain the inherited information in Fig \ref{fig:moti}:
\begin{equation}
Z_{Inh} = \lambda R. 
\label{equ:lambda}
\end{equation}

Finally, $X_{Info}$ is formulated by:
\begin{equation}
{X}_{Info}=\mathcal{D}(LoRA(Z_{Inh}))
\end{equation}

\subsubsection{Training loss}
\label{section:loss}

Training losses contain data-level reconstruction loss and manifold-level information loss.

As for data loss, the image-level supervision for $X_{Info}$ is formulated as MSE loss:
\begin{equation}
\begin{aligned}
\mathcal{L}_{L2} = ||X_{Info}-X_{HQ}||_{2}^{2}.
\end{aligned}
\end{equation}

We add the perceptual loss to improve the high-quality blind face restoration via
\begin{equation}
\mathcal{L}_{vgg}=\lambda_{vgg}^{(i)}\sum ^{N}_{i=1} \| F_{vgg}^{(i)} (X_{Info}) -F_{vgg}^{(i)} ({X}_{HQ}) \| _{2},  
\end{equation}
where $F_{vgg}^{(i)}$ denotes the $i$-th convolution layer of the VGG19 model. We set N equal to 5. $\lambda_{vgg}^{(i)}$ is set to 1/32, 1/16, 1/8, 1/4 and 1.0 in order.

To sum up, we set $\mathcal{L}_{data}$ as the sum of $\mathcal{L}_{L2}$, $\mathcal{L}_{vgg}$ and additional $\mathcal{L}_{LPIPS}(X_{Info};X_{HQ})$ \cite{lpips}:
\begin{equation}
\mathcal{L}_{data}=\mathcal{L}_{L2}+\mathcal{L}_{vgg}+\lambda_{LPIPS}\mathcal{L}_{LPIPS},  
\end{equation}

For Gaussian distribution $\mathcal{N}(\mu, \sigma^{2})$ and $\mathcal{N}(0,1)$, KL divergence is formulated as:
\begin{equation}
\begin{aligned}
\mathcal{D}_{KL}[N(\mu, \sigma^{2})\Vert N(0,1)]
                   = -\frac{1}{2}[log(\sigma)^{2}-(\sigma)^{2}-(\mu)^{2}+1].
\end{aligned}
\end{equation}

As for our case mentioned in Equ. \ref{equ:3}, the distribution of $p(Z|R)$ is accessed as $\mathcal{N}[\lambda R+(1-\lambda)\mu_{\mathcal{M}},(1-\lambda)^{2}\sigma_{\mathcal{M}}^{2}]$ according to Equ. \ref{equ:compress}. We normalize both $p(Z|R)$ and $p(Z)$ using $\mu_{\mathcal{M}}$ and $\sigma_{\mathcal{M}}$, then the information compression metric is formulated as:
\begin{equation}
\begin{aligned}
\mathcal{L}_{info} &= \mathbb{I}(Z;R) \\
                   &= KL[p(Z|R)\Vert q(Z)]\\
                   &= -\frac{1}{2}[log(1-\lambda)^{2}-(1-\lambda)^{2}-(\lambda\frac{R-\mu_{\mathcal{M}}}{\sigma_{\mathcal{M}}})^{2}+1],
\end{aligned}
\end{equation}
based on which KL--divergence controls the distribution distance between original and compressed manifolds. For example, $KL$==0 $\rightarrow p(Z|R)\sim\mathcal{N}(\mu_{\mathcal{M}},\sigma_{\mathcal{M}}^{2})\rightarrow\lambda=0$, which means the percent of information compression of manifold is 100$\%$. InfoBFR finetunes stable diffusion for further controllable face syntheses according to the optimized manifold.

Finally, the total loss of InfoBFR is formulated as:

\begin{equation}
\mathcal{L}_{\emph{InfoBFR}}= \beta\mathcal{L}_{info}+  \mathcal{L}_{data},
\label{con:20}
\end{equation}
where $\beta$ is the trade-off weight to regulate the extent of information compression (Fig. \ref{fig:bt}). 

\section{Experiment}
\label{others}

\subsection{Experimental Protocol}
\subsubsection{Training Protocol} We train our InfoBFR on the FFHQ \cite{style2019} dataset with $512\times512$ size.  We train InfoBFR for 1000k iterations with one NVIDIA RTX 4090 GPU. The training batch size is set to 1. We utilize Stable Diffusion 2.1 as the generative prior with LoRA of rank 4. During training, we employ AdamW \cite{adamw} with $10^{-4}$ learning rate. We set $\lambda_{LPIPS}=2, \beta=20$. We calculate $\mu_{QC}$ and $\sigma_{QC}$ using 3000 HQ samples of FFHQ. Moreover, the manifold feature after QC (quant\_conv layer) has 8 channels with the same setting as $d$ in Equ. \ref{equ:att} and information filter $\lambda$. Note that minimal std $o$ in Algorithm \ref{alg:alg1} is set to 0.01.
\begin{table}[tbp]
\centering
\huge
\caption{Performance boosting study of InfoBFR for state-of-the-art BFR methods in Tab. \ref{tab:real}.  The comprehensive boosting score is calculated through $\frac{1}{N_{dataset}}\sum ^{N_{dataset}}_{i=1}(Score_{BFR}-Score_{BFR^{*}})$ for the first three indexes, and $\frac{1}{N_{dataset}}\sum ^{N_{dataset}}_{i=1}(Score_{BFR^{*}}-Score_{BFR})$ for MUSIQ. Red and blue colors indicate the first and second scores, respectively. NIQE and MUSIQ indicators may have declined slightly because MIB filters neural noises that seem sharp. Nevertheless, around \textcolor{red}{85\%} evaluation events have obvious BFR-boosting. }
\scalebox{0.4}{
\begin{tabular}{c|c|c|c|c}
\hline 
 Methods & \makecell[c]{FID} & \makecell[c]{KID} & \makecell[c]{NIQE} & \makecell[c]{MUSIQ}\tabularnewline
\hline 
\hline
 
VQFR* \emph{vs} VQFR \cite{vqfr}&\textcolor{blue}{11.12}	&1.11	&-0.52	&-0.11
\tabularnewline
CodeFormer* \emph{vs} CodeFormer \cite{codeformer}&3.28	&0.1	&-0.12	&-0.39
\tabularnewline
GPEN* \emph{vs} GPEN \cite{GPEN}&8.94	&0.95	&0.45	&-0.89
\tabularnewline
GFPGAN* \emph{vs} GFPGAN \cite{gfp2021}&4.94	&0.43	&-0.19	&2.62
\tabularnewline
DFDNet* \emph{vs} DFDNet \cite{dfdnet2020} &6.50 	&0.23	&0.55	&\textcolor{blue}{6.97}
\tabularnewline
FeMaSR* \emph{vs} FeMaSR \cite{femasr}&\textcolor{red}{38.76}	&\textcolor{red}{2.97} 	&\textcolor{blue}{0.85}	&\textcolor{red}{15.89}
\tabularnewline
DifFace* \emph{vs} DifFace \cite{difface}&7.86	&0.53 	&0.02	&4.25
\tabularnewline
DiffBIR* \emph{vs} DiffBIR \cite{diffbir}&7.1	&0.57	&\textcolor{red}{1.07}	&4.12
\tabularnewline
StableSR* \emph{vs} StableSR \cite{stablesr}&3.59	&0.23	&0.18 	&1.34
\tabularnewline
GCFSR* \emph{vs} GCFSR \cite{GCFSR}&1.55	&\textcolor{blue}{2.89}	&-0.04	&0.92 
\tabularnewline

\hline 
\end{tabular}}

\label{tab:improve}
\end{table}
\begin{table*}[htbp]
\centering
\caption{Quantitative evaluation compared with state-of-the-art methods (VQFR \cite{vqfr}, CodeFormer \cite{codeformer}, GPEN \cite{GPEN}, GFPGAN \cite{gfp2021}, DFDNet \cite{dfdnet2020}, FeMaSR \cite{femasr}, GCFSR \cite{GCFSR}, DifFace \cite{difface}, DiffBIR \cite{diffbir} and StableSR \cite{stablesr}) on Wider-Test, FOS, CelebChild, and WebPhoto datasets. Note that \textbf{$*$} means the pre-trained BFR model improvement equipped with our proposed InfoBFR. Red, blue, and green colors indicate the first, second, and third scores, respectively. Mention that InfoBFR wins \textcolor{red}{\textbf{10}} 1st prizes, \textcolor{blue}{\textbf{12}} 2nd prizes and \textcolor{green}{\textbf{9}} 3rd prizes in \textbf{16} competition events. Real-world performances of WebPhoto and FOS indicate InfoBRF's high real-world generalization. It's surprising that nearly 80\% evaluation events have obvious BFR-boosting and even FeMaSR* has 38.76 FID-boosting comprehensively as shown in Tab. \ref{tab:improve}.}
\scalebox{0.75}{
\begin{tabular}{c|cccc|cccc|cccc|cccc}
\hline 
\makecell[c]{Datasets}  &\multicolumn{4}{c|}{Wider-Test \cite{codeformer}} &\multicolumn{4}{c|}{FOS \cite{fos}}& \multicolumn{4}{c|}{CelebChild \cite{gfp2021}} & \multicolumn{4}{c}{WebPhoto \cite{gfp2021}} \tabularnewline
\hline 
\makecell[c]{Methods} & \makecell[c]{FID$\downarrow$} &  \makecell[c]{\tiny KID×100\\ \tiny± std.×100}$\downarrow$ & \makecell[c]{NIQE$\downarrow$} & \makecell[c]{MUSIQ$\uparrow$}& \makecell[c]{FID$\downarrow$} &  \makecell[c]{\tiny KID×100\\ \tiny ± std.×100}$\downarrow$ & \makecell[c]{NIQE$\downarrow$} & \makecell[c]{MUSIQ$\uparrow$}& \makecell[c]{FID$\downarrow$} & \makecell[c]{\tiny KID×100\\ \tiny ± std.×100}$\downarrow$ & \makecell[c]{NIQE$\downarrow$} & \makecell[c]{MUSIQ$\uparrow$}& \makecell[c]{FID$\downarrow$} & \makecell[c]{\tiny KID×100\\ \tiny ± std.×100}$\downarrow$ & \makecell[c]{NIQE$\downarrow$} & \makecell[c]{MUSIQ$\uparrow$}\tabularnewline
\hline 
\hline
VQFR \cite{vqfr} &62.96	&\makecell[c]{3.81\\ $\pm$0.09}	&\textcolor{red}{3.44}	&72.02&
105.18	&\makecell[c]{2.03\\ $\pm$0.18}	&\textcolor{red}{4.24}	&68.97 &102.27	&\textcolor{green}{\makecell[c]{3.06\\ $\pm$0.28}}	&\textcolor{red}{4.55}	&\textcolor{green}{70.92}	&80.6	&\makecell[c]{3.56\\ $\pm$0.20}	&\textcolor{red}{4.38}	&71.35	\tabularnewline

VQFR* &40.09	&\makecell[c]{1.10\\ $\pm$0.06}	&\textcolor{blue}{4.01}	&72.54&
92.93	&\textcolor{green}{\makecell[c]{1.07\\ $\pm$0.19}}	&4.82	&68.92 &\textcolor{red}{99.35}	&\textcolor{blue}{\makecell[c]{2.78\\ $\pm$0.26}}	&\textcolor{blue}{4.75}	&69.97	&\textcolor{red}{74.17}	&\textcolor{blue}{\makecell[c]{3.06\\ $\pm$0.24}}	&5.14	&71.37	\tabularnewline
\hline

CodeFormer \cite{codeformer} &44.1	&\makecell[c]{1.68\\ $\pm$0.09}	&4.73	&72.36 &96.38	&\makecell[c]{1.41\\ $\pm$0.17}	&\textcolor{blue}{4.42}	&69.87 &104.19	&\makecell[c]{3.59\\ $\pm$0.34}	&5.69	&70.83	&78.87	&\textcolor{green}{\makecell[c]{3.16\\ $\pm$0.18}}	&5.59	&70.51	\tabularnewline

CodeFormer* &\textcolor{blue}{35.55}	&\textcolor{green}{\makecell[c]{0.86\\ $\pm$0.05}}	&4.6	&72.07&
93.43	&\makecell[c]{1.30\\ $\pm$0.16}	&5.65	&67.51 &103.5	&\makecell[c]{3.60\\ $\pm$0.33}	&5.17	&70.74	&77.96	&\makecell[c]{3.69\\ $\pm$0.19}	&5.52	&71.65	\tabularnewline
\hline

GPEN \cite{GPEN}&49.59	&\makecell[c]{2.38\\ $\pm$0.09}	&4.56	&\textcolor{green}{72.70} &99.38	&\makecell[c]{1.99\\ $\pm$0.17}	&4.88	&\textcolor{blue}{71.80}  &121.32	&\makecell[c]{5.69\\ $\pm$0.28}	&6.45	&\textcolor{blue}{71.42}	&97.49	&\makecell[c]{6.16\\ $\pm$0.21}	&6.65	&\textcolor{red}{73.41}	\tabularnewline

GPEN* &43.04	&\makecell[c]{1.78\\ $\pm$0.09}	&4.78	&72.62&
92.35	&\makecell[c]{1.52\\ $\pm$0.19}	&5.00 	&\textcolor{green}{70.84} &108.43	&\makecell[c]{4.06\\ $\pm$0.27}	&5.28	&69.95	&88.20 	&\makecell[c]{5.04\\ $\pm$0.22}	&5.71	&72.33	\tabularnewline
\hline

GFPGAN \cite{gfp2021}&47.86	&\makecell[c]{2.21\\ $\pm$0.08}	&\textcolor{green}{4.26}	&70.95&
94.73	&\makecell[c]{1.43\\ $\pm$0.14}	&\textcolor{green}{4.65}	&70.11&	107.24	&\makecell[c]{3.60\\ $\pm$0.31}	&4.97	&70.58	&84.14	&\makecell[c]{4.45\\ $\pm$0.31}	&\textcolor{blue}{4.85}	&68.04	\tabularnewline
GFPGAN* &37.97	&\makecell[c]{1.17\\ $\pm$0.06}	&4.45	&\textcolor{red}{72.95}&
\textcolor{green}{90.74}	&\makecell[c]{1.11\\ $\pm$0.15}	&4.81	&\textcolor{red}{71.90} &103.23	&\makecell[c]{3.50\\ $\pm$0.34}	&5.13	&\textcolor{red}{72.38}	&82.27	&\makecell[c]{4.19\\ $\pm$0.24}	&\textcolor{green}{5.12}	&\textcolor{blue}{72.93}	  \tabularnewline
\hline

DFDNet \cite{dfdnet2020}&57.18	&\makecell[c]{2.73\\ $\pm$0.07}	&6.33	&64.73&
101.39	&\makecell[c]{2.16\\ $\pm$0.13}	&5.64	& 57.52 &\textcolor{blue}{101.92}	&\textcolor{red}{\makecell[c]{2.62\\ $\pm$0.22}}	&\textcolor{green}{4.75}	&69.44	&87.97	&\makecell[c]{4.29\\ $\pm$0.19}	&6.57	&63.81	\tabularnewline
DFDNet* &43.04	&\makecell[c]{1.76\\ $\pm$0.10}	&5.00 	&72.07&
93.17	&\makecell[c]{1.54\\ $\pm$0.16}	&5.25	&68.69 &102.44	&\makecell[c]{3.32\\ $\pm$0.24}	&5.05	&70.42	&83.82	&\makecell[c]{4.27\\ $\pm$0.18}	&5.78	&72.21	 \tabularnewline
\hline
FeMaSR \cite{femasr}&115.95	&\makecell[c]{8.01\\ $\pm$0.09}	&6.25	&41.63&
137.14	&\makecell[c]{4.38\\ $\pm$0.18}	&6.49	&46.57&110.72	&\makecell[c]{3.09\\ $\pm$0.19}	&6.45	&66.54	&99.69	&\makecell[c]{4.42\\ $\pm$0.16}	&5.93	&57.09	\tabularnewline
FeMaSR* &37.92	&\textcolor{blue}{\makecell[c]{0.78\\ $\pm$0.05}}	&5.33	&67.76&
92.13	&\makecell[c]{1.15\\ $\pm$0.14}	&5.31	&67.30  &103.63	&\makecell[c]{3.10\\ $\pm$0.29}	&5.40 	&69.69	&\textcolor{blue}{74.77}	&\textcolor{red}{\makecell[c]{3.00\\ $\pm$0.16}}	&5.68	&70.65	\tabularnewline
\hline
GCFSR \cite{GCFSR}&38.87	&\makecell[c]{1.24\\ $\pm$0.06}	&5.67	&69.69&
93.30 	&\makecell[c]{1.23\\ $\pm$0.13}	&6.11	&66.44 &107.74	&\makecell[c]{3.39\\ $\pm$0.27}	&6.03	&67.42	&85.43	&\makecell[c]{4.41\\ $\pm$0.20}	&6.57	&69.28	\tabularnewline
GCFSR* &38.61	&\makecell[c]{1.31\\ $\pm$0.08}	&4.71	&72.43&
90.85	&\makecell[c]{1.19\\ $\pm$0.17}	&5.12	&70.04 &106.55	&\makecell[c]{3.69\\ $\pm$0.36}	&5.21	&69.99	&83.14	&\makecell[c]{4.27\\ $\pm$0.19}	&5.68	&71.93	\tabularnewline
\hline
DifFace \cite{difface}&44.21	&\makecell[c]{1.82\\ $\pm$0.08}	&4.88	&67.14&
104.51	&\makecell[c]{1.76\\ $\pm$0.18}	&4.87	&64.40&105.76	&\makecell[c]{3.54\\ $\pm$0.34}	&5.21	&68.04	&87.31	&\makecell[c]{4.25\\ $\pm$0.20}	&5.36	&67.87	 \tabularnewline
DifFace* &\textcolor{green}{36.96}	&\makecell[c]{1.16\\ $\pm$0.06}	&4.8	&72.18&
\textcolor{blue}{90.16}	&\textcolor{red}{\makecell[c]{0.90\\ $\pm$0.13}}	&5.00	&70.11&105.69	&\makecell[c]{3.73\\ $\pm$0.34}	&5.11	&70.01	&\textcolor{green}{77.52}	&\makecell[c]{3.49\\ $\pm$0.21}	&5.35	&72.12	\tabularnewline
\hline

DiffBIR \cite{diffbir}&38.54	&\makecell[c]{1.04\\ $\pm$0.04}	&5.98	&68.37&
99.92	&\makecell[c]{1.47\\ $\pm$0.16}	&6.21	&64.47&110.17	&\makecell[c]{3.90\\ $\pm$0.32}	&5.95	&68.26	&89.04	&\makecell[c]{4.75\\ $\pm$0.22}	&6.78	&67.23	\tabularnewline
DiffBIR* &\textcolor{red}{34.85}	&\textcolor{red}{\makecell[c]{0.72\\ $\pm$0.05}}	&4.72	&72.29&
\textcolor{red}{90.07}	&\textcolor{blue}{\makecell[c]{1.00\\ $\pm$0.11}}	&5.13	&70.23&104.57	&\makecell[c]{3.33\\ $\pm$0.29}	&5.17	&70.48	&79.8	&\makecell[c]{3.84\\ $\pm$0.21}	&5.61	&71.81	\tabularnewline
\hline
StableSR \cite{stablesr}&40.76	&\makecell[c]{1.25\\ $\pm$0.05}	&4.86	&71.43&
98.18	&\makecell[c]{1.68\\ $\pm$0.11}	&4.88	&67.30 &106.79	&\makecell[c]{3.88\\ $\pm$0.30}	&5.49	&70.16	&79.94	&\makecell[c]{3.66\\ $\pm$0.18}	&5.74	&72.00	\tabularnewline
StableSR* &37.47	&\makecell[c]{1.12\\ $\pm$0.06}	&4.58	&\textcolor{blue}{72.79}&
91.95 &\makecell[c]{1.25\\ $\pm$0.15} &5.09 &69.93&\textcolor{green}{102.03} &\makecell[c]{3.38\\ $\pm$0.24} &5.12	&70.90 &79.87 &\makecell[c]{3.81\\ $\pm$0.19} &5.45	&\textcolor{green}{72.61}  \tabularnewline
\hline

\end{tabular}
}
\label{tab:real}
\end{table*}
\begin{figure*}[htbp]
\begin{center}
   \includegraphics[width=0.9\linewidth]{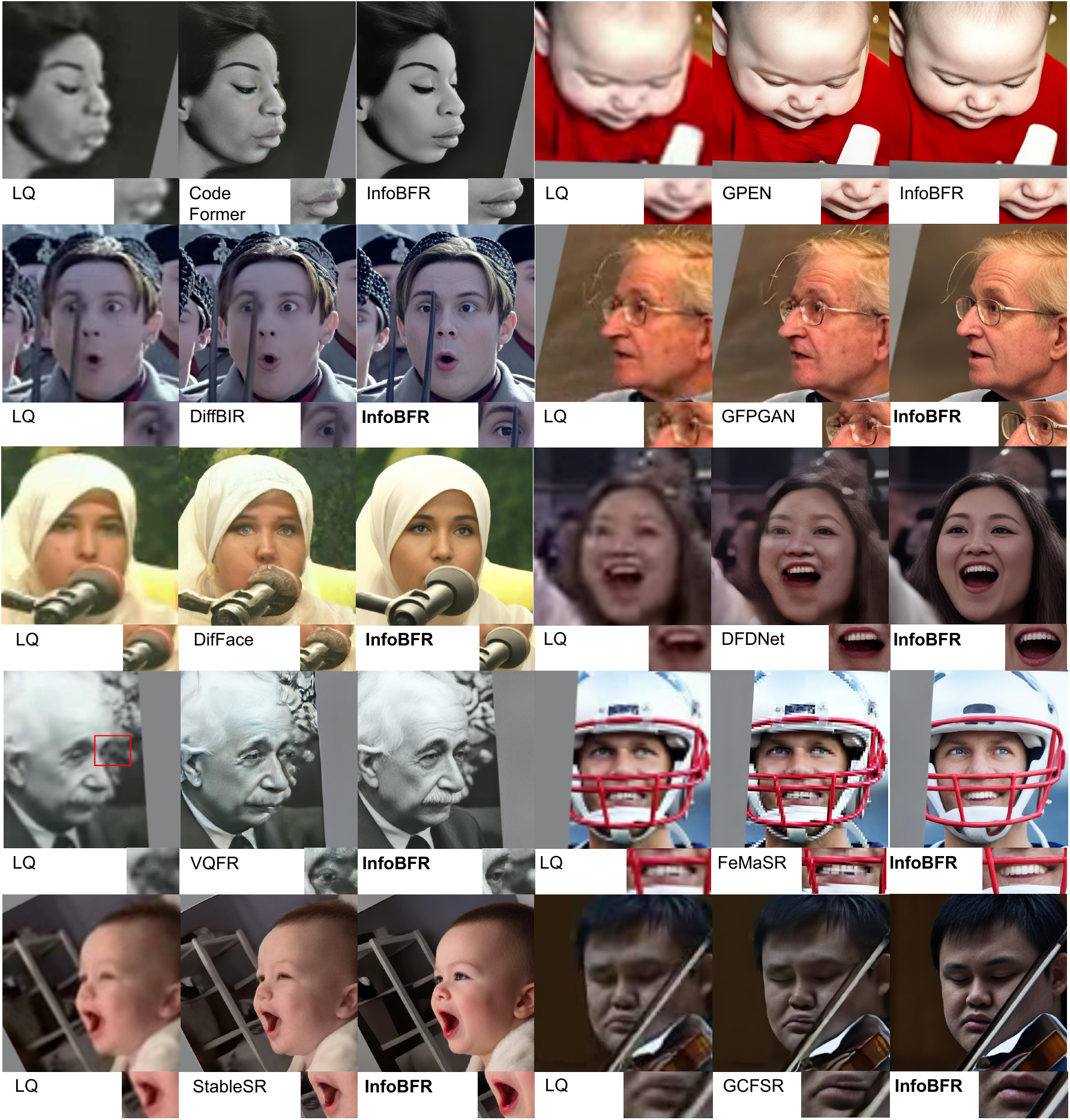}
\end{center}
   \caption{Qualitative comparison with state-of-the-art methods (CodeFormer \cite{codeformer}, GPEN \cite{GPEN}, DiffBIR \cite{diffbir}, GFPGAN \cite{gfp2021}, DifFace \cite{difface}, DFDNet \cite{dfdnet2020},  VQFR \cite{vqfr},  FeMaSR \cite{femasr},  StableSR \cite{stablesr} and GCFSR \cite{GCFSR}) in severely degraded photorealistic scenes including side, large-pose, occluded, extreme-expression faces. InfoBFR is capable of higher fidelity and more realistic face restoration. Zoom in for better observation.}
\label{fig:extreme}
\end{figure*}

\begin{figure*}[htbp]
\begin{center}
   \includegraphics[width=1\linewidth]{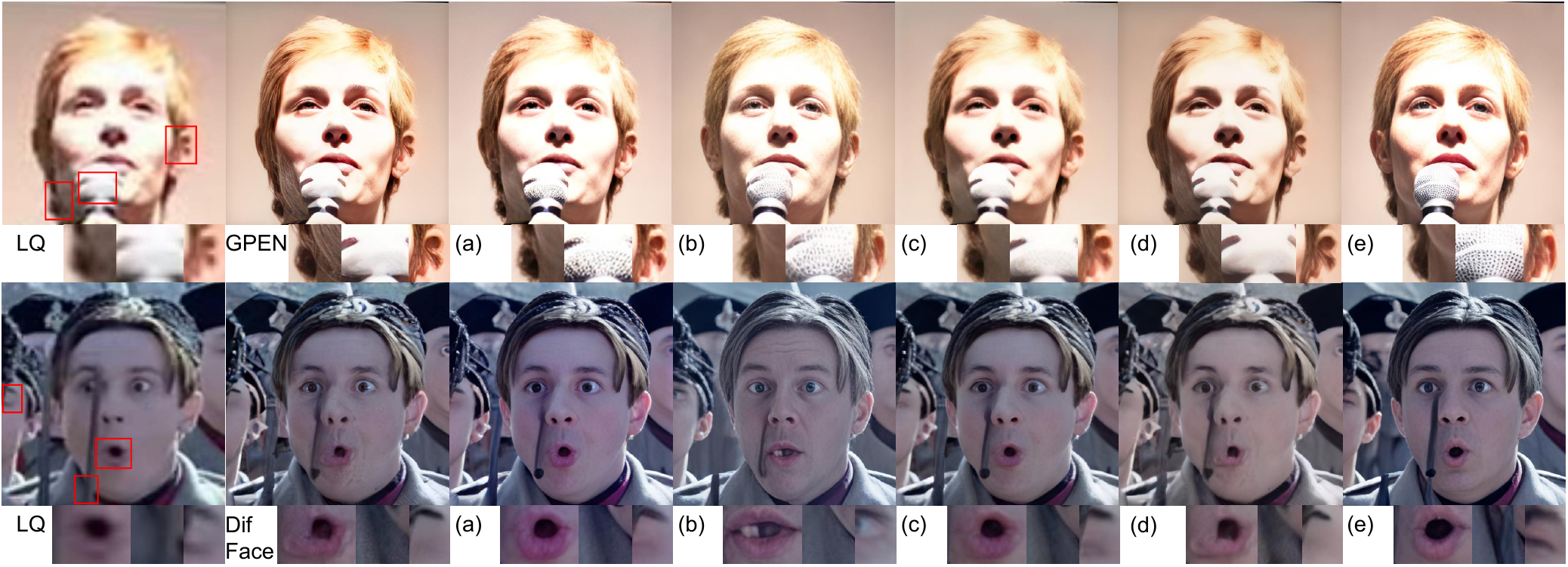}
\end{center}
   \caption{Ablation study concerning different InfoBIR settings, i.e., (a). w/ Transformer, w/o MIB, w/ LoRA, (b). w/o Transformer, w/ MIB,  w/ LoRA, (c). w/o Transformer, w/o MIB,  w/ LoRA, (d). w/ Transformer, w/ MIB, w/o LoRA, (e). w/ Transformer, w/ MIB, w/ LoRA. (e) successfully carries on pleasant and high-fidelity restoration against neural degradations, e.g., topology distortion and textural distortion with facial occlusions, demonstrating the functional importance of Transformer and MIB. More quantitative results are shown in Tab \ref{tab:ablation}.}
\label{fig:ablation} 
\end{figure*}
\begin{figure}[htbp]
\begin{center}
   \includegraphics[width=1\linewidth]{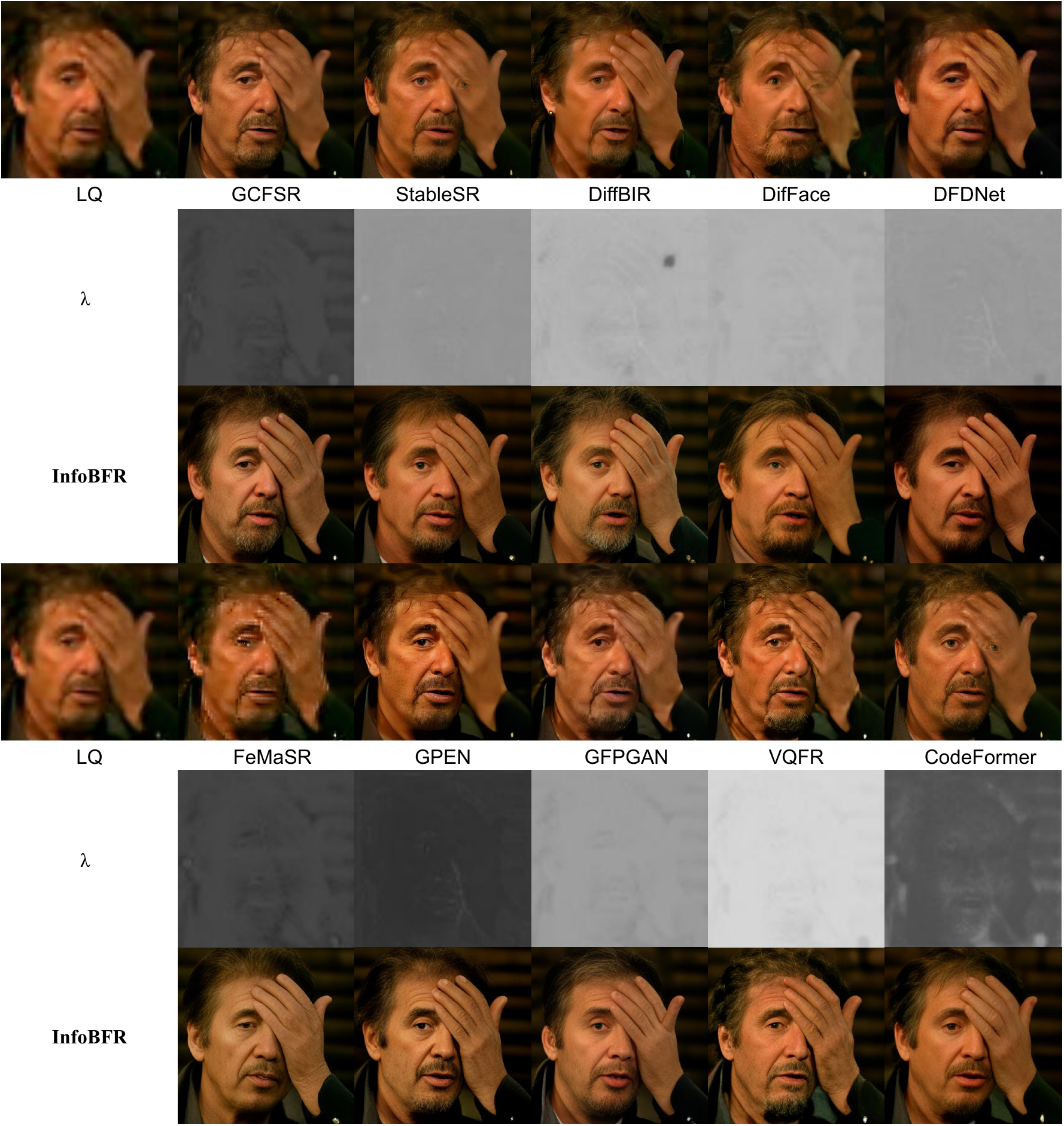}
\end{center}
   \caption{Compressed manifold mask ($\lambda$ in Equ. \ref{equ:lambda}) visualization of diverse BFR models.  InfoBFR adaptively conducts information compression based on MIB and effectively conducts appealing and true-to-life restoration, maintaining facial occlusions. Nevertheless, original BFR models without InforBFR result in undesired structure and texture distortion.}
\label{fig:att} 
\end{figure}

we apply the degradation model adopted in \cite{codeformer}\cite{diffbir} to synthesize the training data with a similar distribution to the real LQ images with data degradations. The degradation pipeline is as follows:  
\begin{equation}
X_{LQ}=\{[(X_{HQ}\otimes \mathbf{k}_{\mathcal{G}})_{\downarrow r}+\mathbf{n}_{\sigma}]_{JPEG_{q}}\}_{\uparrow r},
\end{equation}
where $\mathbf{k}_{\mathcal{G}}$ denotes Gaussian blur kernel with $\mathcal{G} \in \{0.1:12\}$. Moreover, the down-sampling scale r, Gaussian noise $\mathbf{n}_{\sigma}$, and JPEG compression quality $q$ are in the range of $\{0.8 : 8\}$, $\{0 : 20\}$, and $\{30 : 100\}$, respectively. These degradation operations are randomly integrated into the training stage.

\subsubsection{Test Datasets} We evaluate the performance of our InfoBFR framework on the real-world datasets with heavy data degradations, i.e., Wider-Test \cite{codeformer}, CelebChild \cite{gfp2021}, WebPhoto \cite{gfp2021} and FOS \cite{fos}. It is worth highlighting that real-world datasets are more challenging for current BFR models. They have more obvious neural degradations and require more assistance from InfoBFR.

\begin{itemize}


\item \textbf{Wider-Test \cite{codeformer}} is a challenging real-world dataset with severely degraded face images collected from the WIDER Face dataset \cite{wider-test}. More visually improved results of InfoBFR are shown in Fig. 1 of Supp.
 
\item \textbf{FOS\cite{fos}} consists of 158 side or occluded faces in complex real-world scenarios for comprehensive BFR evaluation (Fig. \ref{fig:extreme}).

\item \textbf{CelebChild\cite{gfp2021}} consists of 180 real-world child pictures in color or grayscale space.

\item \textbf{WebPhoto\cite{gfp2021}} consists of 407 LQ faces collected on the Internet. Most of them are from Asian faces.

\end{itemize}
\vspace{-5.pt}
\subsubsection{Evaluation Metric}

We compare the performance of our method with state-of-the-art. The metrics include FID \cite{FID}, Kernel Inception Distance (KID×100 ± std.×100) \cite{kid}, luminance-level NIQE \cite{NIQE} and transformer-based MUSIQ \cite{musiq}. Tab. \ref{tab:real} shows a quantitative comparison of the real-world dataset. Note that the measure anchor of FID and KID is based on FFHQ.

KID \cite{kid} possesses an unbiased estimator, which endows it with enhanced reliability, particularly in scenarios where the dimensionality of the inception features is more than the number of test images \cite{ugatit2020,nicegan}. NIQE \cite{NIQE} and MUSIQ \cite{musiq} are non-reference blind image quality assessment methods. Furthermore, we conduct the performance boosting study of InfoBFR for state-of-the-art BFR methods in Tab. \ref{tab:improve}.

\subsubsection{Comparison methods} 
\label{section:method}
We evaluate the performance of our InfoBFR framework with 8 recent state-of-the-art methods, i.e., dictionary-based methods (CodeFormer \cite{codeformer}, VQFR \cite{vqfr}, FeMaSR \cite{femasr}, DFDNet \cite{dfdnet2020}), StyleGAN-based methods (GPEN \cite{GPEN}, GFPGAN \cite{gfp2021}), diffusion-based methods (DifFace \cite{difface}, DiffBIR \cite{diffbir}, StableSR \cite{stablesr}), and a non-prior method GCFSR \cite{GCFSR}. Here we briefly summarise their advantages and limitations.

\begin{itemize}
\item \textbf{CodeFormer \cite{codeformer}} models the global face composition with codebook-level contextual attention, which implements stable BFR within the representation scope of the learned codebook. It is liable to fall into facial prior overfitting (col 2 in Fig \ref{fig:start}), as well as the mismatch between predicted code and proposed topology of the LQ input (col 1 in Fig \ref{fig:start}).

\item \textbf{VQFR \cite{vqfr}}  is equipped with a VQ codebook dictionary and a parallel decoder for high-fidelity BFR with HQ textural details. It synthesizes sharper images with abundant textural details usually demonstrated by better NIQE \cite{NIQE} score (Tab. \ref{tab:real}). However, these detail enhancements are not conducted based on useful-information screening, which may have a serious impact on facial topology and global fidelity (row 4 in Fig. \ref{fig:stage2}, Fig. \ref{fig:extreme}).

\item \textbf{DifFace \cite{difface}}  builds a Markov chain partially on the pre-trained diffusion reversion for BFR. By denoting the observed LQ image and restored image as $y_{0}$ and $x_{0}$, the intrinsic formulation is to approximate $q(x_{N}|x_{0})$ via $p(x_{N}|y_{0})$ where its diffused estimator and the fake $x_{N}$ may result in important information missing of original LQ face and uncontrollable diffusion reversion, illustrated by the distortions of occlusion (row 3 in Fig. \ref{fig:extreme}), facial components (row 3 in Fig. 3 of Supp.).  
\begin{table}[htbp]
\centering
\huge
\caption{Ablation study of different comparison factors on FOS \cite{fos}. MIB and LoRA are crucial tools for enhancing the performance of the algorithm. More visual results are shown in Fig. \ref{fig:ablation}. Moreover, we show the comparison among InfoBFR with different trade-off weights, i.e., $\beta$ in Equ. \ref{equ:mib}. Note that (b), (d), and (e) adopt $\beta$=20. More visual results are shown in Fig. \ref{fig:bt}.}
\scalebox{0.38}{
\begin{tabular}{c|c c c|c c}
\hline 
 Exp. & Transformer & MIB & LoRA & FID$\downarrow$ & NIQE$\downarrow$ \tabularnewline
\hline 
\hline
GPEN &— 	& —	& —&99.38	&\textcolor{red}{4.88}\tabularnewline
GPEN$*$ (a) &\usym{1F5F8} 	& 	&\usym{1F5F8}&93.78	&5.84 \tabularnewline

(b) & 	&\usym{1F5F8} 	&\usym{1F5F8}&\textcolor{red}{91.17}	&5.10 \tabularnewline

(c) & 	& 	&\usym{1F5F8}&93.75	&5.53 \tabularnewline
(d) &\usym{1F5F8} 	&\usym{1F5F8} 	&	&101.28 &8.19 \tabularnewline
(e) &\usym{1F5F8} 	&\usym{1F5F8} 	&\usym{1F5F8} &\textcolor{blue}{92.35}	&\textcolor{blue}{5.00}\tabularnewline
\hdashline
(e-$\beta_{1}$) &\usym{1F5F8} 	&\usym{1F5F8} 	&\usym{1F5F8} &92.89 &5.48\tabularnewline
(e-$\beta_{5}$) &\usym{1F5F8} 	&\usym{1F5F8} 	&\usym{1F5F8} &93.58 &5.00\tabularnewline
(e-$\beta_{10}$) &\usym{1F5F8} 	&\usym{1F5F8} 	&\usym{1F5F8} &94.10 &5.34\tabularnewline
\hdashline
(e-rank2) &\usym{1F5F8} 	&\usym{1F5F8} 	&\usym{1F5F8} &91.36 &5.15\tabularnewline
(e-rank8) &\usym{1F5F8} 	&\usym{1F5F8} 	&\usym{1F5F8} &92.94 &5.09\tabularnewline
\hline 
DifFace &— 	&—	&— &104.51	&\textcolor{red}{4.87} \tabularnewline
DifFace$*$ (a) &\usym{1F5F8} 	& 	&\usym{1F5F8}&90.46	& 5.85\tabularnewline

(b) & 	&\usym{1F5F8} 	&\usym{1F5F8}&\textcolor{red}{86.70}	&5.15 \tabularnewline

(c) & 	& 	&\usym{1F5F8}&90.20	&5.58 \tabularnewline
(d) &\usym{1F5F8} 	&\usym{1F5F8} 	& &107.97	&7.76 \tabularnewline
(e) &\usym{1F5F8} 	&\usym{1F5F8} 	&\usym{1F5F8}&\textcolor{blue}{90.16}	&\textcolor{blue}{5.00}\tabularnewline
\hdashline
(e-$\beta_{1}$) &\usym{1F5F8} 	&\usym{1F5F8} 	&\usym{1F5F8} &94.36 &5.70\tabularnewline
(e-$\beta_{5}$) &\usym{1F5F8} 	&\usym{1F5F8} 	&\usym{1F5F8} &92.58 &5.46\tabularnewline
(e-$\beta_{10}$) &\usym{1F5F8} 	&\usym{1F5F8} 	&\usym{1F5F8} &94.31 &6.04 \tabularnewline
\hdashline
(e-rank2) &\usym{1F5F8} 	&\usym{1F5F8} 	&\usym{1F5F8} &88.65 &5.46\tabularnewline
(e-rank8) &\usym{1F5F8} 	&\usym{1F5F8} 	&\usym{1F5F8} &90.74 &5.33\tabularnewline
\hline 
\end{tabular}}

\label{tab:ablation}
\end{table}
\begin{figure*}[htbp]
\begin{center}
   \includegraphics[width=0.9\linewidth]{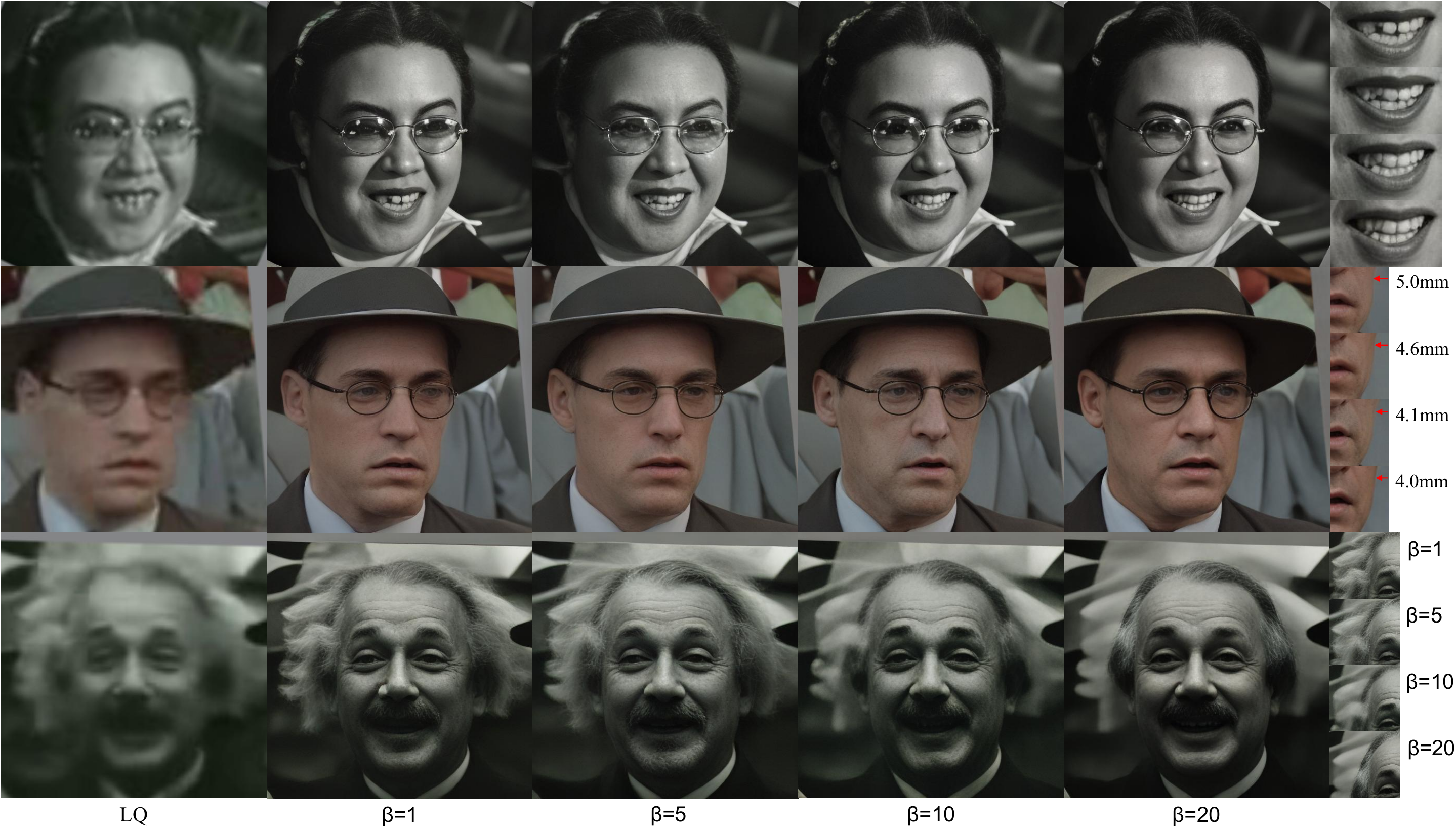}
\end{center}
   \caption{Qualitative results for different $\beta$ in Equ. \ref{equ:mib}. If the value of $\beta$ is larger, more information is compressed, and the generated face better conforms to the real-world face distribution. However, this may result in the slight loss of some detailed attributes of the original character.}
\label{fig:bt} 
\end{figure*}
\begin{table*}[tbp]
\centering
\huge
\caption{Time study of InfoBFR and other state-of-the-art methods on NVIDIA RTX 4090. DifFace \cite{difface}, DiffBIR \cite{diffbir} and  StableSR \cite{stablesr} take more time based on the diffusion model. InforBFR exhibits competitive performance in terms of time consumption by virtue of one-step diffusion LoRA.}
\scalebox{0.4}{
\begin{tabular}{c|c|c|c|c|c|c|c|c|c|c|c}
\hline 
 Methods & \makecell[c]{Code \\Former} & VQFR  & DifFace&FeMaSR& GPEN&StableSR & GFPGAN& DFDNet & DiffBIR& GCFSR & InfoBFR \tabularnewline
\hline 
\hline
 
Sec&0.02	&0.11	&2.48	&0.03	&0.01&8.55&0.02	&0.44	&2.65&0.01 &0.07\tabularnewline

\hline 
 
Trainable Param (M)&376.6 &306.4 &639.4 &177.0 &284.1 &150.0 &615.4 &961.7 &380.0 &355.0 &16.4\tabularnewline
\hline
\end{tabular}}

\label{tab:time}
\end{table*}

\item \textbf{FeMaSR \cite{femasr}} leverages the same learning paradigm as CodeFormer and VQFR, that is high-resolution prior storing along with feature matching. The training is conducted on patches to avoid content bias, which is beneficial to natural image restoration but not BFR with high facial composition prior in global view (row 4 in Fig. \ref{fig:stage2}, Fig. \ref{fig:extreme}).

\item \textbf{GPEN \cite{GPEN}} first trains StyleGAN from scratch with a modified GAN block, then embed multi-level features to the pre-trained prior model with concat operation. Similar to DifFace, GPEN employs latent codes from LQ face as the approximate Gaussian noises to modulate the pre-trained prior model, which is susceptible to intrinsic insufficient information of LQ face. Therefore, details are lacking on the restored face (row 3 in Fig. 2 of Supp.), and even original hallucinations make face reconstruction deviate from the reasonable distribution (row 4 in Fig. 2 of Supp.). 

\item \textbf{GFPGAN \cite{gfp2021}} is another StyleGAN-prior-based method. Similar to GPEN, there is an uncontrollable risk while using the severely abstract latent from LQ face to drive the decoding of pre-trained StyleGAN. Moreover, the naive two-part channel splitting of the CS-SFT layer makes it hard to accurately incorporate realness-aware and fidelity-aware features, which usually results in attribute distortions (row 2 in Fig. \ref{fig:stage2}, Fig. \ref{fig:extreme}).

\item \textbf{DFDNet \cite{dfdnet2020}} conducts multi-level dictionary feature transfer based on HQ dictionaries of facial components. Nonetheless, it is challenging for out-of-component restoration (row 3 in Fig. \ref{fig:stage2}) and precise HQ feature retrieval, especially in complicated degraded scenarios (row 3 in Fig. \ref{fig:extreme}).

\item \textbf{DiffBIR \cite{diffbir}} designs a two-stage framework to realize BFR including restoration module (RM) and generation module (GM) based on SwinIR and stable diffusion respectively. The results preserve hostile neural degradations (topological and textural distortions) deserving to be removed (row 5\&6 in Fig. 2 of Supp.). 

\item \textbf{GCFSR \cite{GCFSR}} generates controllable  super-resolution faces without facial and GAN priors. It's hard for his from-scratch model to deal with BFR under extreme conditions such as large-pose (Fig. 4 in Supp.).

\item \textbf{StableSR \cite{stablesr}} may synthesize suboptimal BFR results in case of severe degradations and highly reliant on the pre-trained BFR model (e.g., CodeFormer \cite{codeformer}) during diffusion sampling. Moreover, it's extremely time-intensive (Tab. \ref{tab:time}).

\item \textbf{InfoBFR} not only leverages the strong generative ability of a stable diffusion model but also adopts information bottleneck to polish the hallucinated BFR model with neural degradations, which is more adept at accommodating a diverse range of BFR scenarios (Fig \ref{fig:stage2}, Fig \ref{fig:extreme} and Supp.). 
\end{itemize}

\subsection{Ablation Study}

We conduct qualitative and quantitative evaluation among different InfoBFR variants, as shown in Tab. \ref{tab:ablation}, Fig. \ref{fig:ablation}, and Fig. \ref{fig:bt}. 
\begin{itemize}
\item $w/ Transformer$ helps our model more strictly respect the original content of $X_{BFR}$. Otherwise, It is challenging for (b) without the Transformer module to defend against severe neural degradations (e.g., mouth distortion) and realize subjectively good BFR results (row 2 in Fig. \ref{fig:ablation}). This highlights the significance of information compression based on spatial attention learning.

\item $w/ MIB$ is capable of filtering stubborn neural degradations (row 1 in Fig. \ref{fig:ablation}) when comparing (a) and (e). (a) fails to restore the shape of the ear (row 1) and eye (row 2). Moreover, MIB is effective in quantitative boosting demonstrated in Tab. \ref{tab:ablation}.

\begin{figure}[tbp]
\begin{center}
   \includegraphics[width=1\linewidth]{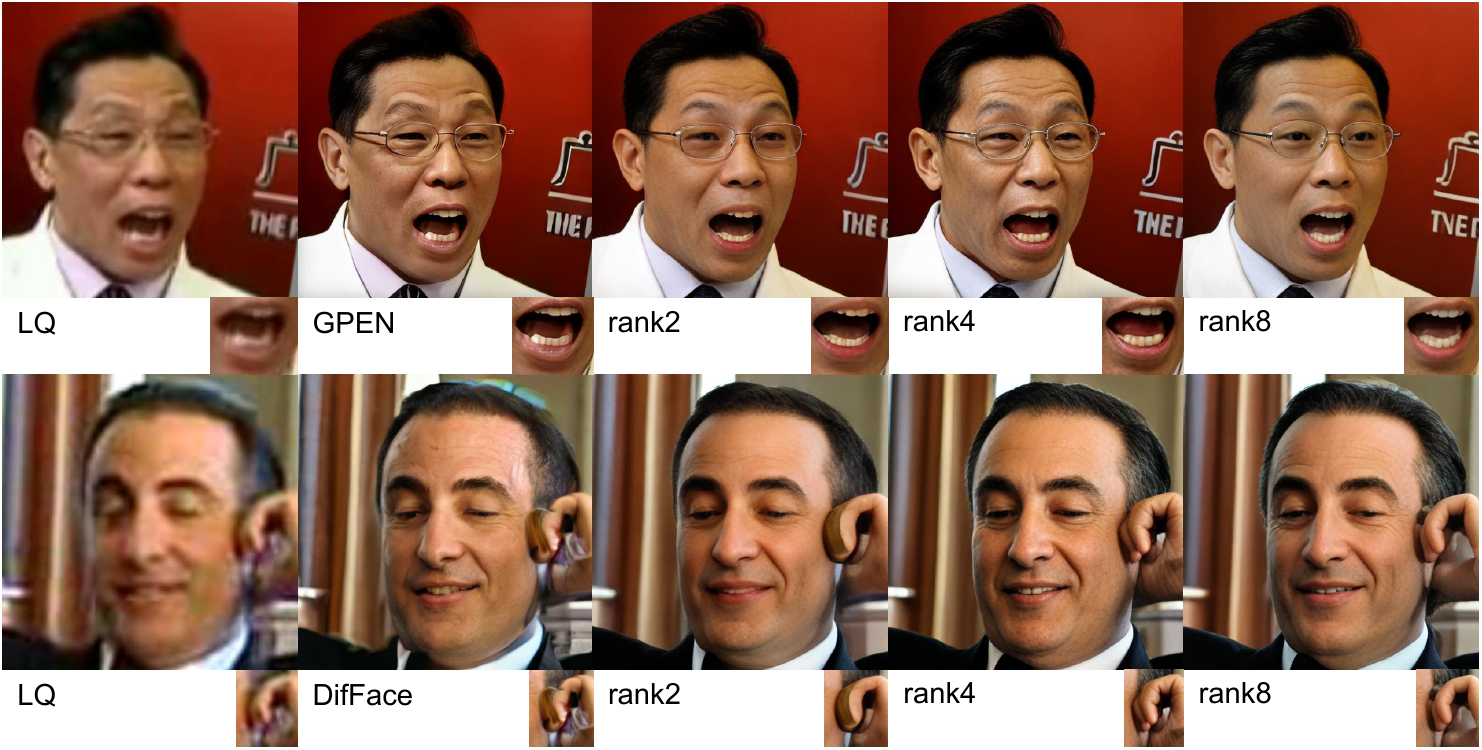}
\end{center}
\vspace{-13.pt}
   \caption{Visual comparative results of InfoBFR with different LoRA rank. More quantitative results are shown in Tab. \ref{tab:ablation} where rank of (e) defaults to 4.}
\label{fig:rank} 
\end{figure}
\item $w/ LoRA$ provides plenty of facial details based on the foundation model. (d) without diffusion LoRA in Tab. \ref{tab:ablation} has the worst FID evaluation that demonstrates the essential role of the LoRA module for information compensation.

\item $\beta$ is friendly to conducting information compression-restoration trade-off. A solid compression weight $\beta$ in Equ. \ref{equ:mib} needs to be explored for stable MIB during InfoBFR training.  Still and all, (e) with $\beta=20$ gets better FID that demonstrates the effectiveness of MIB as shown in Tab. \ref{tab:ablation}. Furthermore, while employing strong compression, InfoBFR is still capable of holding on original identity and expressions due to the constraints from $\mathcal{L}_{data}$.

\item $rank$ controls the information compensation capacity of diffusion LoRA. Note that $rank2$ and $rank8$ have around 8M and 32M trainable parameters, respectively. $rank2$ provides too few facial details, making it difficult to repair the distorted face. There was an overreach for $rank8$ in influencing the generation process of InfoBFR. As shown in Fig \ref{fig:rank}, the texts of (rank8, row1) have been ruined, and the fingers of (rank2, row2) stay far away from the real-world human fingers.

\item \emph{InfoBFR} strictly eliminates the neural degradations caused by pre-trained BFR models and respects the distribution of the real HQ face. This highlights the significance of information optimization based on MIB, Transformer, and diffusion LoRA.

\end{itemize}
\begin{figure}[tbp]
\begin{center}
   \includegraphics[width=1\linewidth]{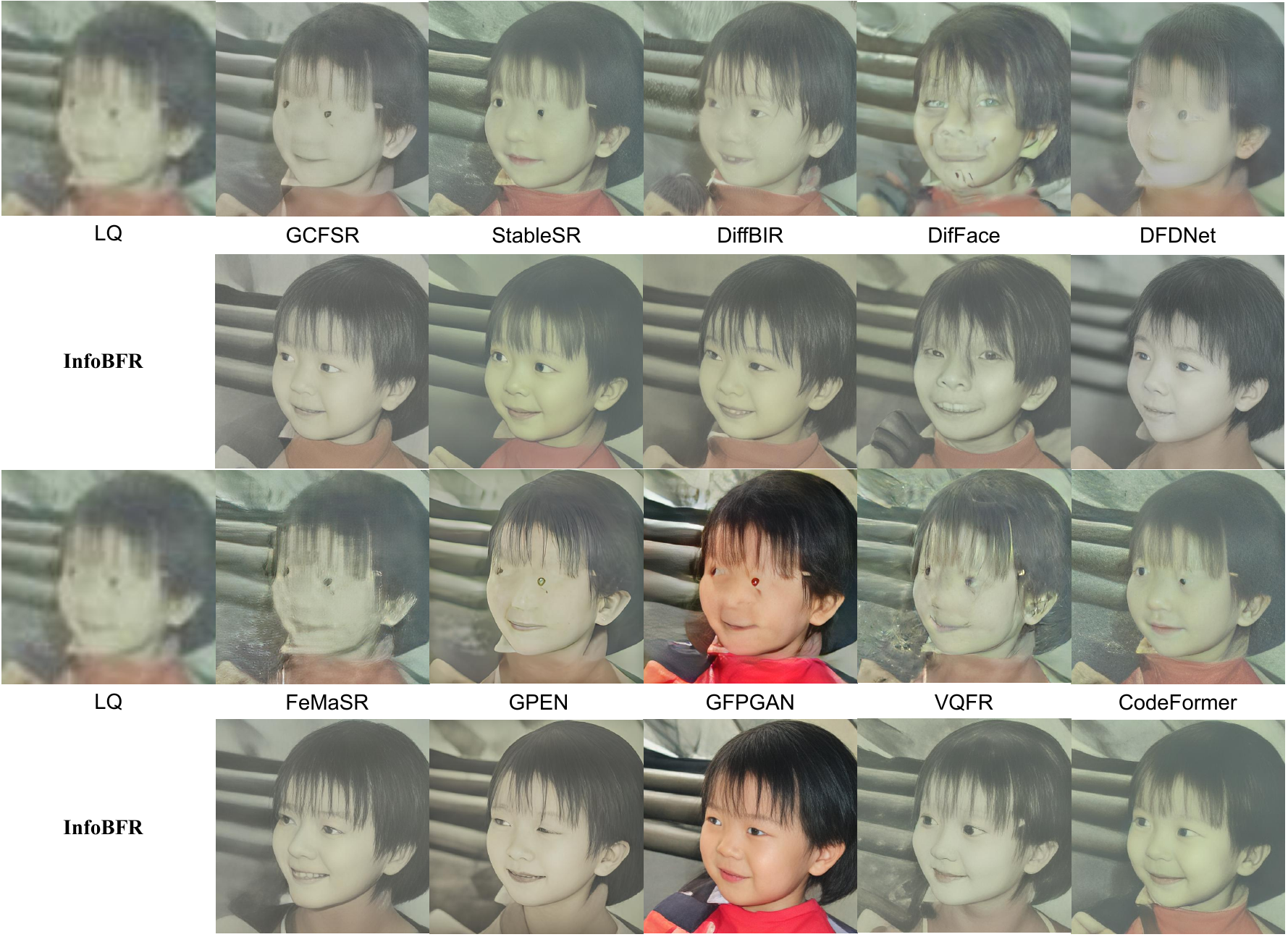}
\end{center}
   \caption{There are some failure generalization cases of InfoBFR in some scenarios, which suffer from severe neural degradations from the pre-trained BFR models. }
\label{fig:hard} 
\end{figure}
\subsection{Discusses}

It is still challenging for InfoBFR to handle BFR in some complicated scenarios with severe neural degradations (Fig \ref{fig:hard}). Nevertheless, InfoBFR has brought about an extremely significant salvation for current BFR models, benefiting from the compression and compensation of information.

As for inference time, InfoBFR takes more inference time compared to some GAN-based models, while it is the best among diffusion-based models (Tab. \ref{tab:time}). 

%


\section{Conclusion}
We first study neural degradations and prose InfoBFR for real-world high-generalization BFR. It is meaningful to explore information restoration based on information compressing and information compensation on manifold space. We propose an effective information optimization framework, including Transformer, MIB, and diffusion LoRA, for high-fidelity and high-quality blind face restoration. Extensive experimental analyses have shown the superiority of InfoBFR compared with state-of-the-art GAN-based and diffusion-based models in complicated wild scenarios, demonstrating that InfoBFR is a plug-and-play boosting method for pre-trained BFR models to conquer neural degradations.


\ifCLASSOPTIONcaptionsoff
  \newpage
\fi

\clearpage
\onecolumn
\centering
\section*{\huge Supplementary material}
\setcounter{figure}{0}
\begin{figure*}[hbp]
\centering
   \includegraphics[width=1\linewidth]{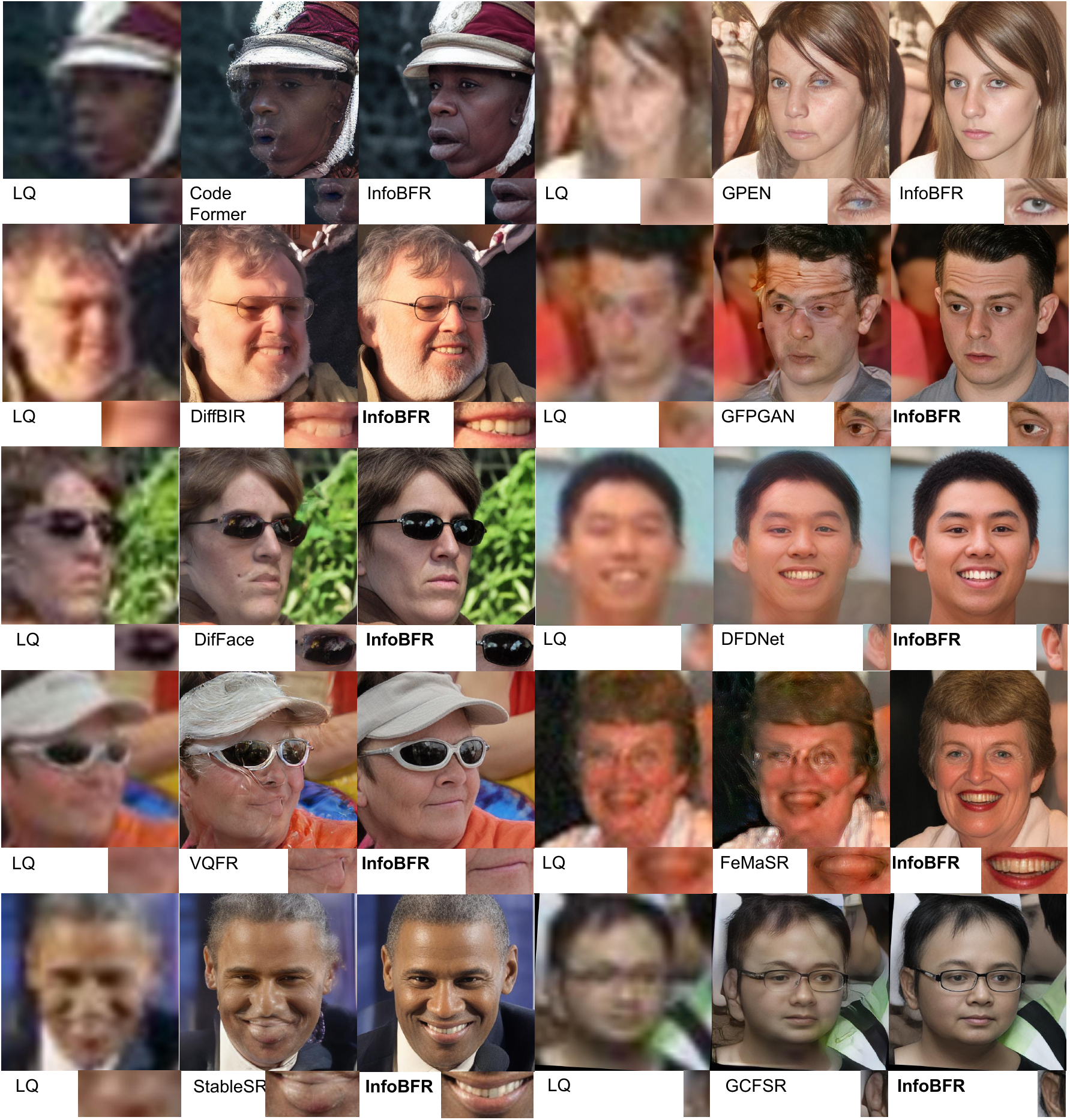}

   \caption{Visual results of state-of-the-art BRF models and the corresponding InfoBFR results in Wider-Test dataset \cite{codeformer}. }
\label{fig:wider}
\end{figure*}

\begin{figure*}[htbp]
\centering
   \includegraphics[width=1\linewidth]{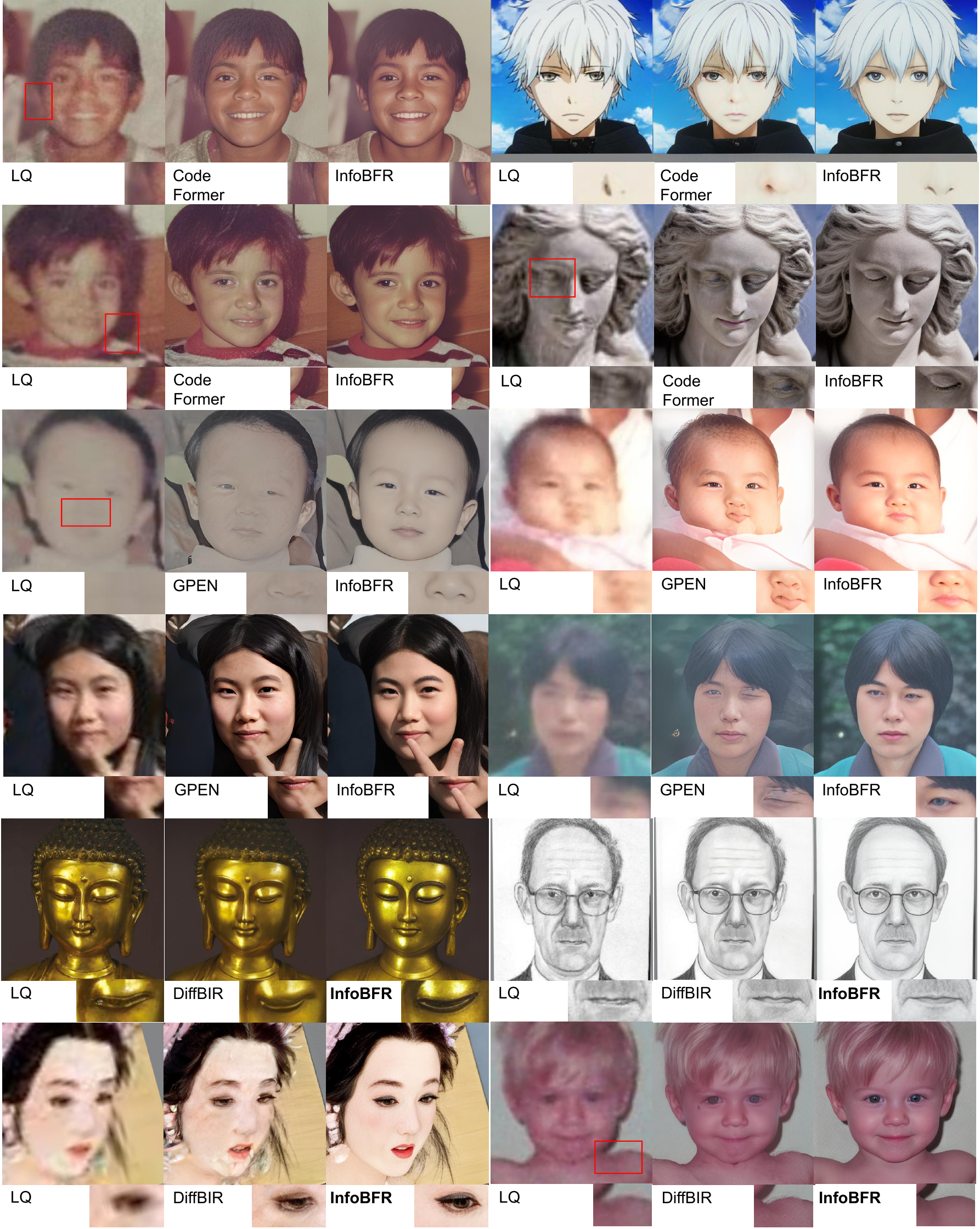}

   \caption{Visual results of CodeFormer \cite{codeformer}, GPEN \cite{GPEN}, DiffBIR \cite{diffbir} and corresponding InfoBFR results. }
\label{fig:gpens}
\end{figure*}

\begin{figure*}[htbp]
\centering
   \includegraphics[width=1\linewidth]{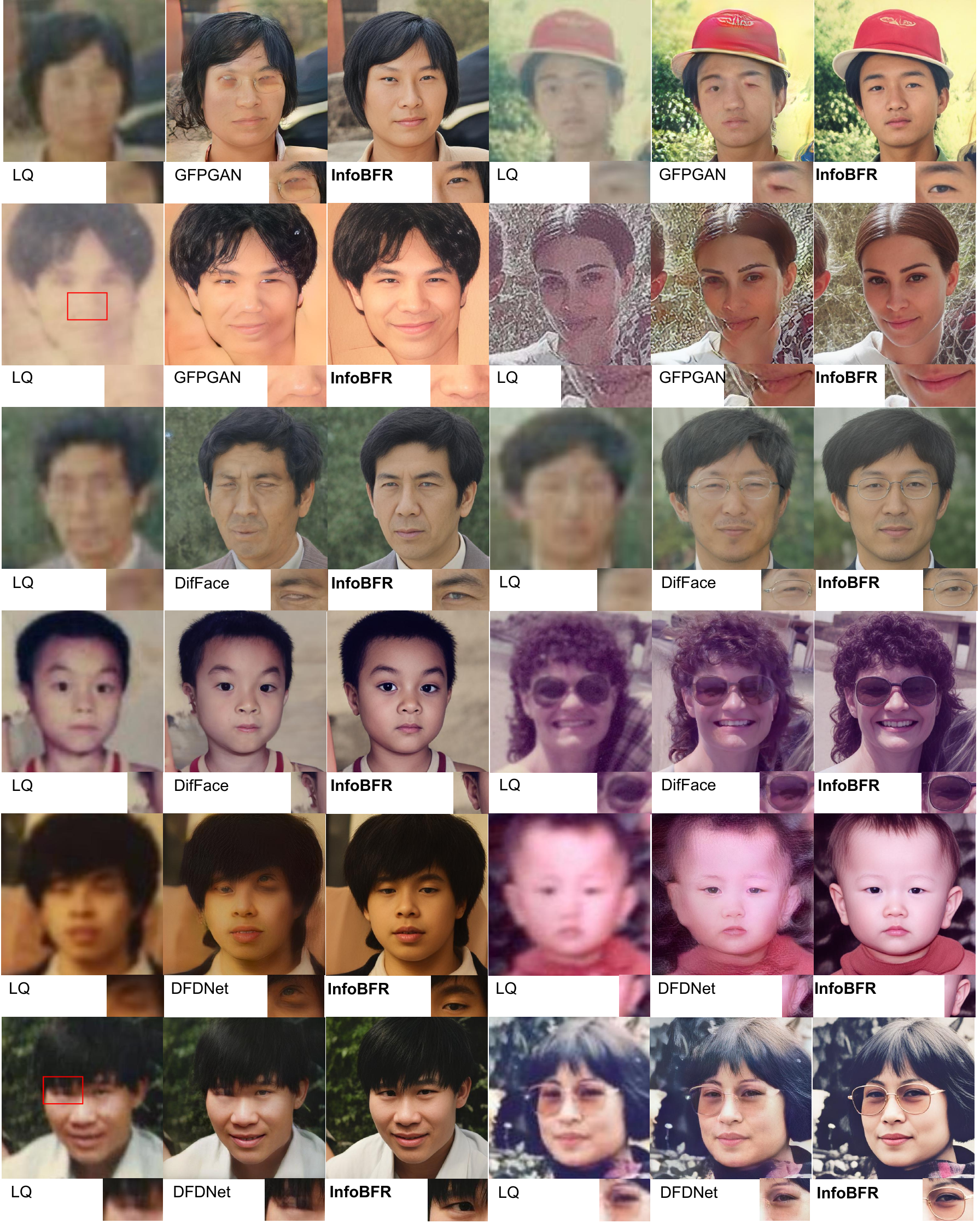}

   \caption{Visual results of GFPGAN \cite{gfp2021}, DifFace \cite{difface}, DFDNet \cite{dfdnet2020} and corresponding InfoBFR results.}
\label{fig:dif}
\end{figure*}
\begin{figure*}[htbp]
\centering
   \includegraphics[width=1\linewidth]{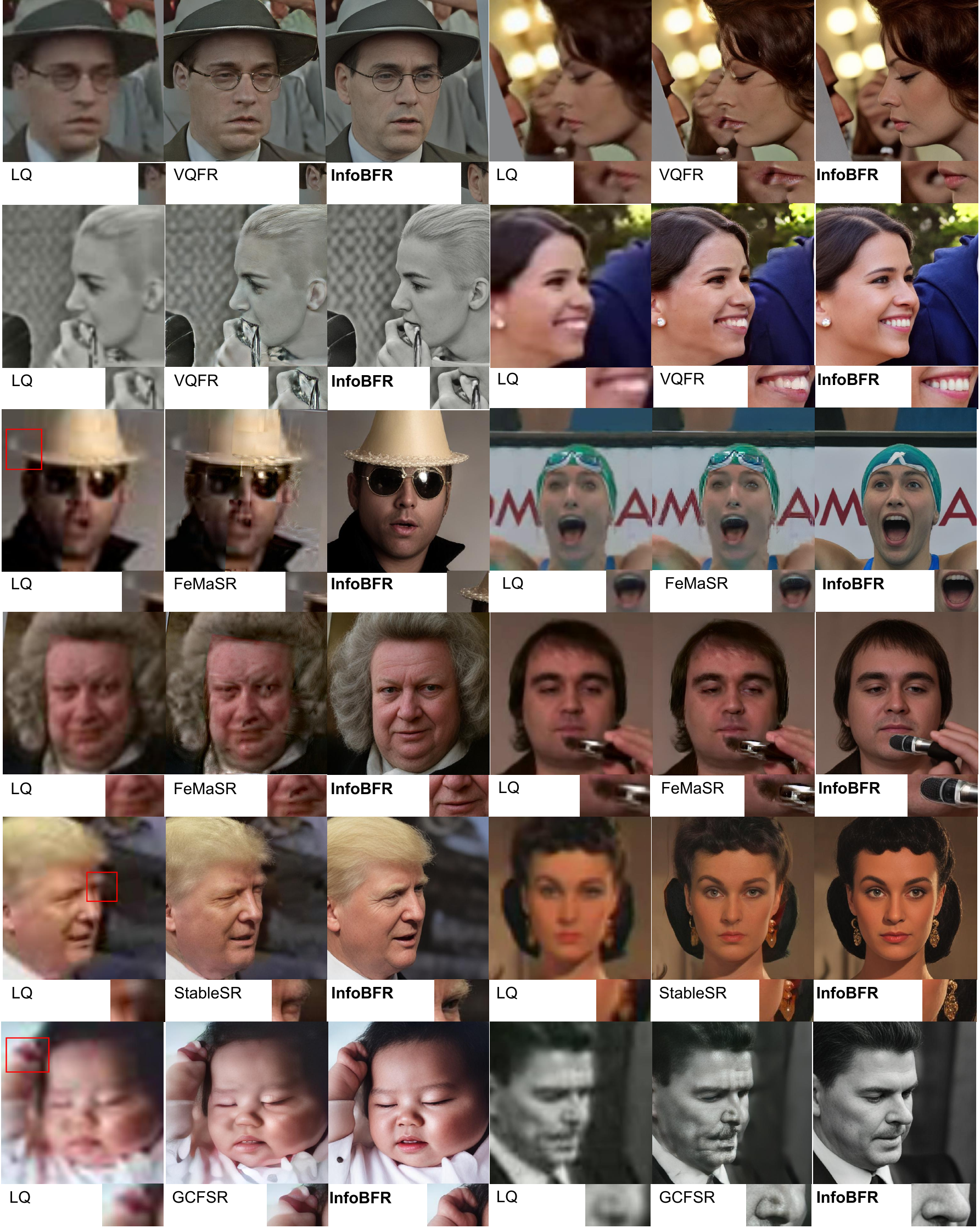}

   \caption{Visual results of VQFR \cite{vqfr}, FeMaSR \cite{femasr}, StableSR \cite{stablesr}, GCFSR \cite{GCFSR} and corresponding InfoBFR results.}
\label{fig:gcfsr}
\end{figure*}


\begin{thebibliography}{1}
\bibitem{iba}
K.~Schulz, L.~Sixt, F.~Tombari, and T.~Landgraf, ``Restricting the flow: Information bottlenecks for attribution,'' in \emph{ICLR}, 2020.


\bibitem{vqgan}
P.~Esser, R.~Rombach, and B.~Ommer, ``Taming transformers for high-resolution image synthesis,'' in \emph{Proceedings of the IEEE/CVF conference on computer vision and pattern recognition}, 2021, pp. 12\,873--12\,883.

\bibitem{vqvae}
A.~Van Den~Oord, O.~Vinyals \emph{et~al.}, ``Neural discrete representation learning,'' \emph{Advances in neural information processing systems}, vol.~30, 2017.

\bibitem{kl}
I.~Csisz{\'a}r, ``I-divergence geometry of probability distributions and minimization problems,'' \emph{The annals of probability}, pp. 146--158, 1975.

\bibitem{ib}
N.~Tishby and N.~Zaslavsky, ``Deep learning and the information bottleneck principle,'' in \emph{2015 ieee information theory workshop (itw)}.\hskip 1em plus 0.5em minus 0.4em\relax IEEE, 2015, pp. 1--5.

\bibitem{vib}
A.~A. Alemi, I.~Fischer, J.~V. Dillon, and K.~Murphy, ``Deep variational information bottleneck,'' \emph{ICLR}, 2017.

\bibitem{trick}
D.~P. Kingma and M.~Welling, ``Auto-encoding variational bayes,'' \emph{ICLR}, 2014.

\bibitem{GCFSR}
J.~He, W.~Shi, K.~Chen, L.~Fu, and C.~Dong, ``Gcfsr: a generative and controllable face super resolution method without facial and gan priors,'' in \emph{2022 IEEE/CVF Conference on Computer Vision and Pattern Recognition (CVPR)}, 2022, pp. 1879--1888.

\bibitem{stablesr}
J.~Wang, Z.~Yue, S.~Zhou, K.~C. Chan, and C.~C. Loy, ``Exploiting diffusion prior for real-world image super-resolution,'' \emph{International Journal of Computer Vision}, pp. 1--21, 2024.

\bibitem{infopaint}
B.~Yang, S.~Gu, B.~Zhang, T.~Zhang, X.~Chen, X.~Sun, D.~Chen, and F.~Wen, ``Paint by example: Exemplar-based image editing with diffusion models,'' in \emph{Proceedings of the IEEE/CVF Conference on CVPR}, 2023, pp. 18\,381--18\,391.




\bibitem{warpnet2018}
X.~Li, M.~Liu, Y.~Ye, W.~Zuo, L.~Lin, and R.~Yang, ``Learning warped guidance
  for blind face restoration,'' in \emph{ECCV}, 2018, pp. 272--289.

\bibitem{exemplar2020}
X.~Li, W.~Li, D.~Ren, H.~Zhang, M.~Wang, and W.~Zuo, ``Enhanced blind face
  restoration with multi-exemplar images and adaptive spatial feature fusion,''
  in \emph{CVPR}, 2020, pp. 2706--2715.

\bibitem{pool2020}
X.~Yan, S.~Cui, ``Towards content-independent multi-reference super-resolution:
  Adaptive pattern matching and feature aggregation,'' in \emph{ECCV}, 2020, pp. 52-68.

\bibitem{copy2020}
Y.~Zhang, I.~W. Tsang, Y.~Luo, C.-H. Hu, X.~Lu, and X.~Yu, ``Copy and paste
  gan: Face hallucination from shaded thumbnails,'' in \emph{CVPR}, 2020, pp.
  7355--7364.

\bibitem{masa2021}
X.~T. J. L. J.~J. Liying~Lu1, Wenbo~Li1, ``Masa-sr: Matching acceleration and
  spatial adaptation for reference-based image super-resolution,'' in
  \emph{CVPR}, 2021.

\bibitem{dfdnet2020}
X.~Li, C.~Chen, S.~Zhou, and et~al., ``Blind face restoration via deep
  multi-scale component dictionaries,'' in \emph{ECCV}.\hskip 1em plus 0.5em
  minus 0.4em\relax Springer, 2020, pp. 399--415.

\bibitem{pulse2020}
S.~Menon, A.~Damian, S.~Hu, and et~al., ``Pulse: Self-supervised photo
  upsampling via latent space exploration of generative models,'' in
  \emph{CVPR}, 2020, pp. 2437--2445.

\bibitem{bank2021}
K.~C. Chan, X.~Wang, X.~Xu, and et~al., ``Glean: Generative latent bank for
  large-factor image super-resolution,'' \emph{arXiv preprint
  arXiv:2012.00739}, 2020.

\bibitem{gfp2021}
X.~Wang, Y.~Li, H.~Zhang, and Y.~Shan, ``Towards real-world blind face
  restoration with generative facial prior,'' in \emph{The IEEE Conference on
  CVPR}, 2021.

\bibitem{GPEN}
T.~Yang, P.~Ren, X.~Xie, and L.~Zhang, ``Gan prior embedded network for blind
  face restoration in the wild,'' in \emph{Proceedings of the IEEE/CVF
  Conference on CVPR}, 2021, pp. 672--681.


\bibitem{style2019}
T.~Karras, S.~Laine, and T.~Aila, ``A style-based generator architecture for
  generative adversarial networks,'' in \emph{CVPR}, 2019, pp. 4401--4410.

\bibitem{stylegan2_2020}
T.~Karras, S.~Laine, M.~Aittala, J.~Hellsten, J.~Lehtinen, and T.~Aila,
  ``Analyzing and improving the image quality of stylegan,'' in \emph{CVPR},
  2020, pp. 8110--8119.



\bibitem{psfr}
C.~Chen, X.~Li, L.~Yang, and et~al., ``Progressive semantic-aware style
  transformation for blind face restoration,'' \emph{arXiv preprint
  arXiv:2009.08709}, 2020.





\bibitem{lfw}
G.~B. Huang, M.~Mattar, T.~Berg, and E.~Learned-Miller, ``Labeled faces in the
  wild: A database forstudying face recognition in unconstrained
  environments,'' in \emph{Workshop on faces in'Real-Life'Images: detection,
  alignment, and recognition}, 2008.


\bibitem{FID}
M.~Heusel, H.~Ramsauer, T.~Unterthiner, B.~Nessler, and S.~Hochreiter, ``Gans
  trained by a two time-scale update rule converge to a local nash
  equilibrium,'' in \emph{NIPS}, 2017.

\bibitem{musiq}
J.~Ke, Q.~Wang, Y.~Wang, P.~Milanfar, and F.~Yang, ``Musiq: Multi-scale image quality transformer,'' in \emph{ICCV}, 2021, pp. 5148--5157.

\bibitem{NIQE}
A.~Mittal, R.~Soundararajan, and A.~C. Bovik, ``Making a “completely blind”
  image quality analyzer,'' \emph{IEEE Signal processing letters}, vol.~20,
  no.~3, pp. 209--212, 2012.

\bibitem{lpips}
R.~Zhang, P.~Isola, A.~A. Efros, E.~Shechtman, and O.~Wang, ``The unreasonable
  effectiveness of deep features as a perceptual metric,'' in \emph{CVPR},
  2018, pp. 586--595.

\bibitem{ugatit2020}
J.~Kim, M.~Kim, H.~Kang, and K.~Lee, ``U-gat-it: unsupervised generative
  attentional networks with adaptive layer-instance normalization for
  image-to-image translation,'' in \emph{ICLR}, 2020.

\bibitem{codeformer}
S.~Zhou, K.~C. Chan, C.~Li, and C.~C. Loy, ``Towards robust blind face
  restoration with codebook lookup transformer,'' in \emph{NeurIPS}, 2022.
  
\bibitem{vqfr}
Y.~Gu, X.~Wang, L.~Xie, C.~Dong, G.~Li, Y.~Shan, and M.-M. Cheng, ``Vqfr: Blind face restoration with vector-quantized dictionary and parallel decoder,'' in \emph{ECCV}, 2022.

\bibitem{infoswap}
G.~Gao, H.~Huang, C.~Fu, Z.~Li, and R.~He, ``Information bottleneck
  disentanglement for identity swapping,'' in \emph{Proceedings of the IEEE/CVF
  conference on CVPR}, 2021, pp. 3404--3413.


\bibitem{controlnet}
L.~Zhang, A.~Rao, and M.~Agrawala, ``Adding conditional control to
  text-to-image diffusion models,'' in \emph{ICCV}, 2023, pp. 3836--3847.

\bibitem{latentdiff}
R.~Rombach, A.~Blattmann, D.~Lorenz, P.~Esser, and B.~Ommer, ``High-resolution
  image synthesis with latent diffusion models,'' in \emph{CVPR}, 2022, pp.
  10\,684--10\,695.


\bibitem{difface}
Z.~Yue and C.~C. Loy, ``Difface: Blind face restoration with diffused error contraction,'' \emph{IEEE Transactions on Pattern Analysis and Machine Intelligence}, 2024.

\bibitem{diffbir}
X.~Lin, J.~He, Z.~Chen, Z.~Lyu, B.~Dai, F.~Yu, Y.~Qiao, W.~Ouyang, and C.~Dong, ``Diffbir: Toward blind image restoration with generative diffusion prior,'' in \emph{European Conference on Computer Vision}.\hskip 1em plus 0.5em minus 0.4em\relax Springer, 2024, pp. 430--448.


\bibitem{femasr}
C.~Chen, X.~Shi, Y.~Qin, X.~Li, X.~Han, T.~Yang, and S.~Guo, ``Real-world blind
  super-resolution via feature matching with implicit high-resolution priors,''
  in \emph{Proceedings of the 30th ACM International Conference on Multimedia},
  2022, pp. 1329--1338.

\bibitem{fos}
Z.~Chen, J.~He, X.~Lin, Y.~Qiao, and C.~Dong, ``Towards real-world video face restoration: A new benchmark,'' in \emph{Proceedings of the IEEE/CVF Conference on Computer Vision and Pattern Recognition (CVPR) Workshops}, June 2024, pp. 5929--5939.

\bibitem{adamw}
I.~Loshchilov and F.~Hutter, ``Decoupled weight decay regularization,'' in \emph{ICLR}, 2019.

\bibitem{ddpm}
A.~Q. Nichol and P.~Dhariwal, ``Improved denoising diffusion probabilistic models,'' in \emph{International Conference on Machine Learning}.\hskip 1em plus 0.5em minus 0.4em\relax PMLR, 2021, pp. 8162--8171.

\bibitem{lora}
E.~J. Hu, Y.~Shen, P.~Wallis, Z.~Allen-Zhu, Y.~Li, S.~Wang, L.~Wang, and W.~Chen, ``Lora: Low-rank adaptation of large language models,'' \emph{arXiv preprint arXiv:2106.09685}, 2021.

\bibitem{denoising}
J.~Ho, A.~Jain, and P.~Abbeel, ``Denoising diffusion probabilistic models,'' \emph{Advances in neural information processing systems}, vol.~33, pp. 6840--6851, 2020.

\bibitem{diffusion}
P.~Dhariwal and A.~Nichol, ``Diffusion models beat gans on image synthesis,'' \emph{Advances in neural information processing systems}, vol.~34, pp. 8780--8794, 2021.

\bibitem{ib_concept}
N.~Tishby, F.~C. Pereira, and W.~Bialek, ``The information bottleneck method,'' \emph{arXiv preprint physics/0004057}, 2000.

\bibitem{kid}
M.~Bi{\'n}kowski, D.~J. Sutherland, M.~Arbel, and A.~Gretton, ``Demystifying mmd gans,'' in \emph{ICLR}, 2018.

\bibitem{wider-test}
S.~Yang, P.~Luo, C.-C. Loy, and X.~Tang, ``Wider face: A face detection benchmark,'' in \emph{Proceedings of the IEEE conference on computer vision and pattern recognition}, 2016, pp. 5525--5533.

\bibitem{nicegan}
R.~Chen, W.~Huang, B.~Huang, F.~Sun, and B.~Fang, ``Reusing discriminators for encoding: Towards unsupervised image-to-image translation,'' in \emph{Proceedings of the IEEE/CVF conference on computer vision and pattern recognition}, 2020, pp. 8168--8177.



\end{thebibliography}
\end{document}